\documentclass[letterpaper, 10 pt, conference]{ieeeconf}  % Comment this line out if you need a4paper

\IEEEoverridecommandlockouts                              % This command is only needed if 
\usepackage{amssymb}
\usepackage{amstext}
\usepackage{amsmath}
\usepackage{multicol, multirow}
\usepackage{dsfont}
\usepackage{pifont}
\usepackage[dvipsnames]{xcolor}
\usepackage{overpic}
\usepackage{tikz}
\usepackage{soul}
\usepackage{url}
\usepackage{listings}
\usepackage{booktabs} % for professional tables
\usepackage[ruled]{algorithm2e}
\usepackage{cite}

\usepackage{adjustbox}

\title{\LARGE \bf
Loop closure detection using local 3D deep descriptors}

\author{Youjie Zhou$^1$, Yiming Wang$^2$, Fabio Poiesi$^2$, Qi Qin$^1$ and Yi Wan$^1$
\thanks{*This work was supported by the China government (2019JZZY010112) and (2020JMRH0202), and by the SHIELD project, funded by the European Union’s Joint Programming Initiative – Cultural Heritage, Conservation, Protection and Use joint call.}
\thanks{$^{1}$Youjie Zhou, Qi Qin and Yi Wan are with the School of Mechanical Engineering, Shandong University, China.  Yi Wan {\tt <wanyi@sdu.edu.cn>} is the corresponding author.}%
\thanks{$^{2}$Yiming Wang and Fabio Poiesi are with Fondazione Bruno Kessler, Italy {\tt <ywang,poiesi>@fbk.eu}.}
}

\begin{document}

\maketitle
\thispagestyle{empty}
\pagestyle{empty}

%%%%%%%%%%%%%%%%%%%%%%%%%%%%%%%%%%%%%%%%%%%%%%%%%%%
%%%%%%%%%%%%%%%%%%%%%%%%%%%%%%%%%%%%%%%%%%%%%%%%%%%
%%%%%%%%%%%%%%%%%%%%%%%%%%%%%%%%%%%%%%%%%%%%%%%%%%%
\begin{abstract}
We present a simple yet effective method to address loop closure detection in simultaneous localisation and mapping using local 3D deep descriptors (L3Ds).
L3Ds are emerging compact representations of patches extracted from point clouds that are learnt from data using a deep learning algorithm.
We propose a novel overlap measure for loop detection by computing the metric error between points that correspond to mutually-nearest-neighbour descriptors after registering the loop candidate point cloud by its estimated relative pose.
This novel approach enables us to accurately detect loops and estimate six degrees-of-freedom poses in the case of small overlaps.
We compare our L3D-based loop closure approach with recent approaches on LiDAR data and achieve state-of-the-art loop closure detection accuracy.
Additionally, we embed our loop closure approach in RESLAM, a recent edge-based SLAM system, and perform the evaluation on real-world RGBD-TUM and synthetic ICL datasets.
Our approach enables RESLAM to achieve a better localisation accuracy compared to its original loop closure strategy.
Our project page is available at \url{github.com/yiming107/l3d_loop_closure}.
\end{abstract}


%%%%%%%%%%%%%%%%%%%%%%%%%%%%%%%%%%%%%%%%%%%%%%%%%%%
%%%%%%%%%%%%%%%%%%%%%%%%%%%%%%%%%%%%%%%%%%%%%%%%%%%
%%%%%%%%%%%%%%%%%%%%%%%%%%%%%%%%%%%%%%%%%%%%%%%%%%%
\section{Introduction}
\label{sec:intro}

Loop closure aims to recognise already visited places in order to mitigate tracking drifts in simultaneous localisation and mapping (SLAM) \cite{Arshad2021}.
Loop closure is a computational module that typically involves {\em loop detection}, {\em pose estimation} and {\em verification}, which can be executed on-board an autonomous robot and that runs in parallel with other processes, e.g.~feature extraction and tracking \cite{MurArtal2017,Schenk2019}.

Loop closure is triggered when there is enough 2D or 3D overlap between the viewpoint of a current keyframe and of a previous one \cite{Schenk2019}. 
Therefore, how to reliably measure the overlap is key to successfully detect loops.
The overlap can be estimated in 2D by extracting visual representations from keyframe images, e.g., through bags of visual words \cite{GalvezTRO12,MurArtal2017, Glocker2015,Schenk2019}, or in 3D by matching geometric or semantic representations between point clouds of keyframes across time \cite{Steder2011,M2DP2016,chen2021overlapnet,IRIS2021,dube2020segmap}.
These 3D representations can be obtained by aggregating local geometric information \cite{Steder2011} or by globally encoding the whole point cloud into a signature \cite{M2DP2016, chen2021overlapnet, wang2020intensity, IRIS2021}. 
Then, the overlap can be either measured through a suitable distance between global keyframe representations, e.g., using the Hamming distance \cite{IRIS2021}, or inferred via a pre-trained Siamese network \cite{chen2021overlapnet}.
Siamese networks can also be used to directly regress the pose between viewpoints, however a minimum overlap is required to obtain an accurate estimate: to estimate a 1DoF transformation between viewpoints, the minimum overlap should be about 30\% \cite{chen2021overlapnet}, whereas the overlap should be greater than 70\% for a 6DoF transformation \cite{Li2021}.
Although loop closure methods that use global representations can be computationally efficient, the literature about geometric representations shows that they lack generalisation ability across domains \cite{Poiesi2021,Ao2021}.
In the context of deep-learning-based algorithms, this limitation implies that when a new domain is visited, extra effort for data collection, annotation and training is needed, thus hindering the use of loop closure for robotic exploration in unseen environments.
On the other hand, recent research shows that local 3D deep descriptors (L3Ds) have better generalisation abilities across domains than their global counterparts \cite{Poiesi2021,Ao2021}.
This is what motivates us to leverage local geometric information for loop closures, in particular to address the specific problems of loop detection and pose estimation.

In this letter, we present a novel approach to detect loop closures where L3Ds are employed to estimate the poses and to quantify the overlaps between loop candidates. 
The main technical novelty of our approach is an overlap measure that is defined as the Ratio Of Nearest points in both the descriptor and metric spaces, namely \emph{RON}.
Specifically, we quantify the overlap between a point cloud pair through the metric error of the registered points that are mutually nearest neighbours in the descriptor space.
Because L3Ds are inferred by a deep network from randomly sampled points within each point cloud, the relative 6DoF transformation between point clouds can be directly estimated with the RANSAC algorithm \cite{Fischler1981}.
Then, RON is computed as the ratio of corresponding points that have a small metric error.

Our L3D-based loop detection approach outperforms the state-of-the-art methods OverlapNet \cite{chen2021overlapnet} and LiDAR Iris \cite{IRIS2021}, which are validated using the LiDAR point clouds of the KITTI odometry dataset \cite{Geiger2012}.
We use the same evaluation strategy as OverlapNet \cite{chen2021overlapnet}.
Moreover, we show the efficacy of our approach to RGBD SLAM systems by embedding it into RESLAM \cite{Schenk2019}, a recent RGBD edge-based SLAM approach. 
We evaluate the absolute trajectory error \cite{Sturm2012} on the real-world TUM-RGBD \cite{Sturm2012} and synthetic ICL \cite{Saeedi2019} datasets.
Our approach enables RESLAM to detect loops more frequently and to estimate 6DoF transformations more accurately than those estimated with the RESLAMS's original loop closure method.
In summary, our contributions are:
\begin{itemize}
    \item a novel use of L3Ds for the loop closure problem;
    \item an overlap measure based on the ratio of nearest points;
    \item a validation for cross-domain applicability with experiments on both LiDAR and RGBD SLAM systems.
\end{itemize}

%%%%%%%%%%%%%%%%%%%%%%%%%%%%%%%%%%%%%%%%%%%%%%%%%%%
%%%%%%%%%%%%%%%%%%%%%%%%%%%%%%%%%%%%%%%%%%%%%%%%%%%
%%%%%%%%%%%%%%%%%%%%%%%%%%%%%%%%%%%%%%%%%%%%%%%%%%%
\section{Related work}\label{sec:related}

We review loop closure detection methods that are designed for point cloud data.
Readers are referred to \cite{Arshad2021} for a thorough review of these.
Methods for loop detection operating in 3D can be categorised into three groups \cite{summary3Dloop2019}: 
feature-based \cite{magnusson2009automatic,M2DP2016,habich2021have,wang2020intensity,Karl2010,IRIS2021}, 
segmentation-based \cite{segmatch2017,fan2020seed}, 
and learning-based \cite{yin2018locnet,dube2020segmap,chen2021overlapnet,zhu2020gosmatch}.

Feature-based methods are typically focused on building hand-crafted rotation-invariant 3D descriptors, for example based on feature histograms \cite{rusu2009fast} or global shape features \cite{rohling2015fast,IRIS2021}.
The Normal Distribution Transform (NDT) is a method that builds global features from histograms of local shape descriptions, where local shapes can be linear, planar or spherical \cite{magnusson2009automatic}.
Each point cloud is divided into overlapping cells and each cell is classified as a specific shape based on the estimated surface normal.
Rotation invariance is achieved by aligning point clouds with respect to dominant surface orientations.
Unlike NDT, M2DP \cite{M2DP2016} first builds intermediate signatures by projecting point clouds to multiple 2D planes and by generating spatial density distributions for each plane.
Then, a global representation is computed by aggregating the left and right singular vectors of these signatures.
Unlike global descriptors that often encode only the geometric properties of the point cloud, the Intensity Scan Context method \cite{wang2020intensity} encodes both the geometry and the intensity of LIDAR scans.
The geometric consistency is then verified through a RANSAC-based registration using FPFH point features \cite{habich2021have}.
Karl \emph{et al.}~\cite{Karl2010} define 41 features of the full scan to create decision stumps, and learn a binary classifier with Adaboost for loop detection.
The LiDAR Iris method \cite{IRIS2021} is inspired by the human’s iris signature for identification in order to detect loop closures efficiently.
Each LiDAR scan is firstly converted into a binary signature by using a series of LoG-Gabor filtering and thresholding operations. 
Then, the Hamming distance between point cloud pairs is used for detecting loops.
Similarly to \cite{Karl2010} we perform a verification step through a RANSAC-based registration but by leveraging deep-learning based descriptors.
Unlike global methods, we do not compute a global representation of a given point cloud/scan, but we instead match local representations between scans to determine a score that provides us with an indication whether a loop closure has occurred or not.

Segmentation-based methods encode the point cloud into a set of discriminative features to reduce the likelihood of false matches.
SegMatch \cite{segmatch2017} uses two hand-crafted features, i.e.~eigenvalue-based and shape histograms, to describe the semantic elements of a scene and performs point cloud matching by using Random Forest together with a RANSAC-based geometric verification step.
Seed~\cite{fan2020seed} employs handcrafted features that encode the topological information of segmented objects to reduce the noise and resolution effects.
Both SegMatch and Seed use a cluster-all approach to segment the point cloud, which requires the ground plane removal prior to the segmentation.
Our approach does not use neither priors nor high-level semantic representations of a scene, it only relies on low-level geometric representations in the form of deep-learning based descriptors.

Learning-based methods include the LocNet approach \cite{yin2018locnet} that computes a handcrafted rotation-invariant representation of a point cloud in a image-like format, which is then processed by a Siamese network to learn global features for place matching and loop closure detection.
Zaganidis \emph{at al.}~\cite{semanticNDT2019} use semantic information processed by PointNet++ \cite{qi2017pointnet++} together with the NDT-based histogram descriptors for loop closure detection. 
GOSMatch \cite{zhu2020gosmatch} builds a global representation in the form of a histogram-based graph descriptor to encode semantic relationships between objects that are segmented by RangeNet++ \cite{milioto2019rangenet++}.
Instead of performing frame-by-frame loop detection, SegMap \cite{dube2020segmap} segments the scene incrementally as the robot navigates and inputs these segments to a deep network to generate a signature per segment.
Loop detection is then performed based on these segments.
Authors in \cite{chen2021overlapnet} and \cite{Chen2021auro}, present OverlapNet that employs a Siamese deep neural network to exploit different types of information of LiDAR scans, including depth, normals, intensity or remission values, to predict the overlap and relative yaw angle (1DoF) between pairs of 3D scans.
Our approach differs from OverlapNet \cite{chen2021overlapnet} in three aspects:
i) in addition to providing a quantitative indication of the overlap between point clouds, our approach can estimate the 6DoF transformation between point clouds;
ii) our approach is local, thus making it suitable to be used in different domains (sensors and scenes);
iii) we operate on point clouds instead of range images.

Most of the above-mentioned methods only estimate the relative yaw angle between LiDAR scans, mainly because public LiDAR datasets, such as KITTI \cite{Geiger2012} or nuScenes \cite{caesar2020nuscenes}, are captured from road vehicles where scans are often co-planar.
Moreover, learning based methods are typically trained and tested on data belonging to the same domain, e.g.~LiDAR data captured outdoors, which cannot generalise well to other domains without retraining or finetuning them, for example on sparser point clouds that are reconstructed with vision-based SLAM systems \cite{rgbdslam_survey2021}.
Differently, our approach exploits deep local 3D descriptors that are trained with point clouds extracted from RGBD sensors, estimates the 6DoF transformation between a pair of point clouds, and measures the relative overlap, serving for the loop closure detection task with domain gap.

%%%%%%%%%%%%%%%%%%%%%%%%%%%%%%%%%%%%%%%%%%%%%%%%%%%
%%%%%%%%%%%%%%%%%%%%%%%%%%%%%%%%%%%%%%%%%%%%%%%%%%%
%%%%%%%%%%%%%%%%%%%%%%%%%%%%%%%%%%%%%%%%%%%%%%%%%%%
\section{L3D-based Loop Closure}\label{sec:method}

A typical SLAM system involves three modules running in parallel that are tracking, local mapping and global mapping \cite{MurArtal2017,Schenk2019}.
Tracking computes the relative camera motion between consecutive frames. The most informative frames are stored as keyframes.
Local mapping processes every new keyframe to incrementally reconstruct the map of the environment and optimises the reconstruction using a time-shifting window.
Global mapping performs loop closure and relocalisation using pose graph optimisation.
Loop closure aims to address the problems of finding keyframes that can potentially form trajectory loops, estimating the poses between candidate keyframe pairs and verifying that the candidate loop closure is correctly estimated.
It is important to detect and solve loops correctly as they would otherwise worsen the trajectory error.

Our proposed L3D-based approach identifies the candidate loop frames by measuring the overlap amongst local 3D descriptors that are extracted between keyframe pairs, and then confirms the occurrence of a loop through pose estimation and RON computation.
In Sec.~\ref{sec:descriptor} we describe the 3D descriptor extraction, while in Sec.~\ref{sec:loop_det} we describe the algorithms for loop detection and confirmation.

%%%%%%%%%%%%%%%%%%%%%%%%%%%%%%%%%%%%%%%%%%%%%%%%%%%
%%%%%%%%%%%%%%%%%%%%%%%%%%%%%%%%%%%%%%%%%%%%%%%%%%%
%%%%%%%%%%%%%%%%%%%%%%%%%%%%%%%%%%%%%%%%%%%%%%%%%%%
\subsection{3D descriptor extraction}
\label{sec:descriptor}

Given a 3D point and a set of its neighbouring points, namely a patch, a local 3D descriptor is a compact numerical representation of these points \cite{Choy2019a,Gojcic2019,Ao2021,Poiesi2021}.

Let $\mathcal{P}_t = \{ \mathbf{p}_t \} \subset \mathbb{R}^3$ be the point cloud of the current keyframe at frame $t$ that is defined as an unordered set of 3D points, and $\Pi_t = [\mathcal{P}_{t-1}, ..., \mathcal{P}_{0}]$ be an ordered collection of point clouds stored up to frame $t-1$, with $t > 0$.
The size of this collection and the number of points of each point may vary over time and across scenes.
At each $t$, we randomly choose a set of $n_t = n \cdot |\mathcal{P}_t|$ points as centres for our 3D descriptors, where $n \in \mathbb{R}_{(0,1]}$ and $|\cdot|$ is the cardinality of a set.
The subsampled point cloud is defined as $\tilde{\mathcal{P}}_t \subset \mathcal{P}_t$ where $|\tilde{\mathcal{P}}_t| = n_t$.
Let $\mathcal{D}_t = \{\mathbf{d}_t\} \subset \mathbb{R}^d$ be the set of $d$-dimensional descriptors where $|\mathcal{D}_t| = |\tilde{\mathcal{P}}_t|$.

Given a point in $\tilde{\mathcal{P}}_t$, we can extract a patch composed of neighbouring points within a spherical region with a certain radius, and use a deep neural network to encode this patch into a compact descriptor.
Let $\mathcal{X}_t = \{ \mathbf{x}_t \} \subset \mathcal{P}_t$ be the patch extracted from $\mathcal{P}_t$ and 
$\mathbf{d}_t = \Phi_{\Theta}(\mathcal{X}_t)$, where $\Phi_{\Theta}$ is the deep neural network with parameters $\Theta$ that encodes $\mathcal{X}_t$ into the descriptor $\mathbf{d}_t$, such that $\lVert \mathbf{d}_t \rVert = 1$. 
We experimentally select the best-performing deep neural network for the descriptor encoding, more details can be found in Sec.~\ref{sec:descriptor_analysis}.

%%%%%%%%%%%%%%%%%%%%%%%%%%%%%%%%%%%%%%%%%%%%%%%%%%%
%%%%%%%%%%%%%%%%%%%%%%%%%%%%%%%%%%%%%%%%%%%%%%%%%%%
%%%%%%%%%%%%%%%%%%%%%%%%%%%%%%%%%%%%%%%%%%%%%%%%%%%
\subsection{Loop detection and verification}\label{sec:loop_det}

When a loop closure occurs, it is highly likely that a portion of the current keyframe's point cloud overlaps with a portion of another point cloud in previous keyframes.
Except for some occlusions, there should exist corresponding surfaces between overlapping point cloud regions, thus corresponding descriptors.
In general, the higher this overlap is, the higher the likelihood of detecting a loop.
We detect candidate loops through the estimation of the overlap between point cloud pairs in the descriptor space, and confirm the occurrence of a loop by computing (i) the transformation to register a candidate point cloud pair and (ii) the novel overlap measure RON that measures the ratio of nearest points in both the descriptor and metric space. 

Specifically, we determine the overlap region between the two point clouds $\tilde{\mathcal{P}}_t$ and $\tilde{\mathcal{P}}_{t'}$, where $t'<t$, by selecting points whose descriptors, $\mathcal{D}_t$ and $\mathcal{D}_{t'}$, are mutually nearest neighbours (MNN).
The computation of MNN descriptors is efficient and does not require the estimation of the transformation matrix between two viewpoints.
Let $\mathcal{C}_{t,t'}$ be the set of corresponding points defined as
%+++++++++++++++++++++
\begin{equation}
  \begin{aligned}
    \mathcal{C}_{t,t'} &= \{ \{ \mathbf{p}_t \in \tilde{\mathcal{P}}_t, \mathbf{p}_{t'} \in \tilde{\mathcal{P}}_{t'}\} : \\
    & \mathbf{d}_t = \mathrm{NN}(\mathbf{d}_{t'}, \mathcal{D}_t) \land \mathbf{d}_{t'} = \mathrm{NN}(\mathbf{d}_t, \mathcal{D}_{t'}) \},
  \end{aligned}
\end{equation}
%+++++++++++++++++++++
where $\mathrm{NN}(\cdot)$ is the nearest neighbour search based on the L2 norm.
We compute the MNN overlap between the two point clouds as
%+++++++++++++++++++++
\begin{equation}
    o_{t,t'} = \frac{|\mathcal{C}_{t,t'}|}{min\{ |\mathcal{D}_t|, |\mathcal{D}_{t'}|\}},
    \label{eq:mnn}
\end{equation}
%+++++++++++++++++++++
where $o_{t,t'} \in \mathbb{R}_{[0,1]}$.
If the overlap $o_{t,t'}$ is greater than a threshold $\tau_o$, then we deem these keyframes to form a candidate loop.

Points with the corresponding descriptors being mutually neighbours does not necessarily guarantee the points in the metric space to be also close to each other, for example flat regions may be ambiguous.
To verifying the loop, we further register the candidate pairs by estimating the relative 6DoF transformation and estimate RON for the loop confirmation. 

Different methods can be used to register two point clouds \cite{Fischler1981,Zhou2016}.
Without loss of generality, we use RANSAC \cite{Fischler1981} as we found that it performs well in practice in terms of computational efficiency and robustness to noise.
Let $\mathbf{T}_{t,t'} \in \mathbb{R}^{4\times4}$ be the transformation estimated with RANSAC between $\tilde{\mathcal{P}}_t$ and $\tilde{\mathcal{P}}_{t'}$ using $\mathcal{D}_t$ and $\mathcal{D}_{t'}$, respectively.

Lastly, with the registered point clouds, we compute RON as the ratio of MNN points in the descriptor space that are close in the metric space, i.e., whose distance in the metric space is below a certain error. 
Let RON be defined as
%+++++++++++++++++++++
\begin{equation}
    \mathtt{RON}_{t,t'} = \frac{1}{|\mathcal{C}_{t,t'}|} \sum_{\{\mathbf{p}_t,\mathbf{p}_{t'}\} \in \mathcal{C}_{t,t'}} \lVert \mathbf{p}_t - \mathbf{T}_{t,t'} \circ \mathbf{p}_{t'} \rVert_2 < \tau_e,
    \label{eq:ron}
\end{equation}
%+++++++++++++++++++++
where $\lVert \cdot \rVert_2$ is the L2 norm, $\circ$ is the operator that applies $\mathbf{T}_{t,t'}$ to a 3D point and $\tau_e$ is the maximum error between two corresponding points that takes into account the reconstruction noise. 
If $\mathtt{RON}_{t,t'}$ is greater than a threshold $\tau_\rho$, then we confirm the occurrence of a loop. 
The transformation $\mathbf{T}_{t,t'}$ can be fed to the module in charge of solving the pose graph problem to optimise the camera poses in any SLAM system. 
The pseudocode for our loop verification approach is shown in Algorithm \ref{alg:dip_loop}.

%*************************************
\begin{algorithm}[t]
\caption{Open3D-style pseudo-code for L3D-based loop detection.}
\label{alg:dip_loop}
\definecolor{codeblue}{rgb}{0.25,0.5,0.5}
\lstset{
    basicstyle=\fontsize{7.2pt}{7.2pt}\ttfamily\bfseries,
    commentstyle=\fontsize{7.2pt}{7.2pt}\color{codeblue},
    keywordstyle=\fontsize{7.2pt}{7.2pt},
    showstringspaces=false,
    literate={'}{{'}}1,
    breakatwhitespace=false,
    breaklines=true,
    breakindent=1\dimen1,
    xleftmargin=-.2cm,
    xrightmargin=.2cm
}

\begin{lstlisting}[language=python]
# P_t: point cloud of keyframe at t
# P_t': point cloud of keyframe at t' < t
# D_t: descriptors of the points in P_t
# D_t': descriptors of the points in P_t'
# nn(x, y): nearest-neighbour (NN) search in the set y using the query x
# T: transformation to register P_t on P_t'
# tau_e: max distance between corresponding points

def mnn(D_t, D_t'):
    # compute the indices of elements in D_t that are NNs of the queries in D_t'
    inds = [nn(d_t', D_t) for d_t' in D_t']
    # compute the indices of elements in D_t' that are NNs of the queries in D_t
    inds' = [nn(d_t, D_t') for d_t in D_t]
    # compute boolean list w.r.t. inds
    # true entries correspond to mutual NN elements
    c_bools = list(range(len(D_t))) == inds[inds']
    return c_bools, inds'

def loop_detection(P_t, P_t', D_t, D_t', T, tau_e):
    # compute mutual nearest neighbours
    c_bools, inds' = mnn(D_t, D_t')
    # compute overlap
    o = c_bools.sum() / min([len(D_t), len(D_t')])
    # apply transformation
    P_t.transform(T)
    # compute L2 norm between mutual NN points
    dists = norm(P_t.points - P_t'.points[inds'])
    # compute ratio of nearest points (RON)
    ron = (dists[c_bools] < tau_e).sum() / c_bools.sum()
    return o, ron
\end{lstlisting}
\end{algorithm}
%*************************************

\begin{figure}[t]
\begin{center}
  \begin{tabular}{@{}c@{}c@{}c}
  %%%%%%%%%%%%%%%%%%%%%%%%%%
    \begin{overpic}[width=.5\columnwidth]{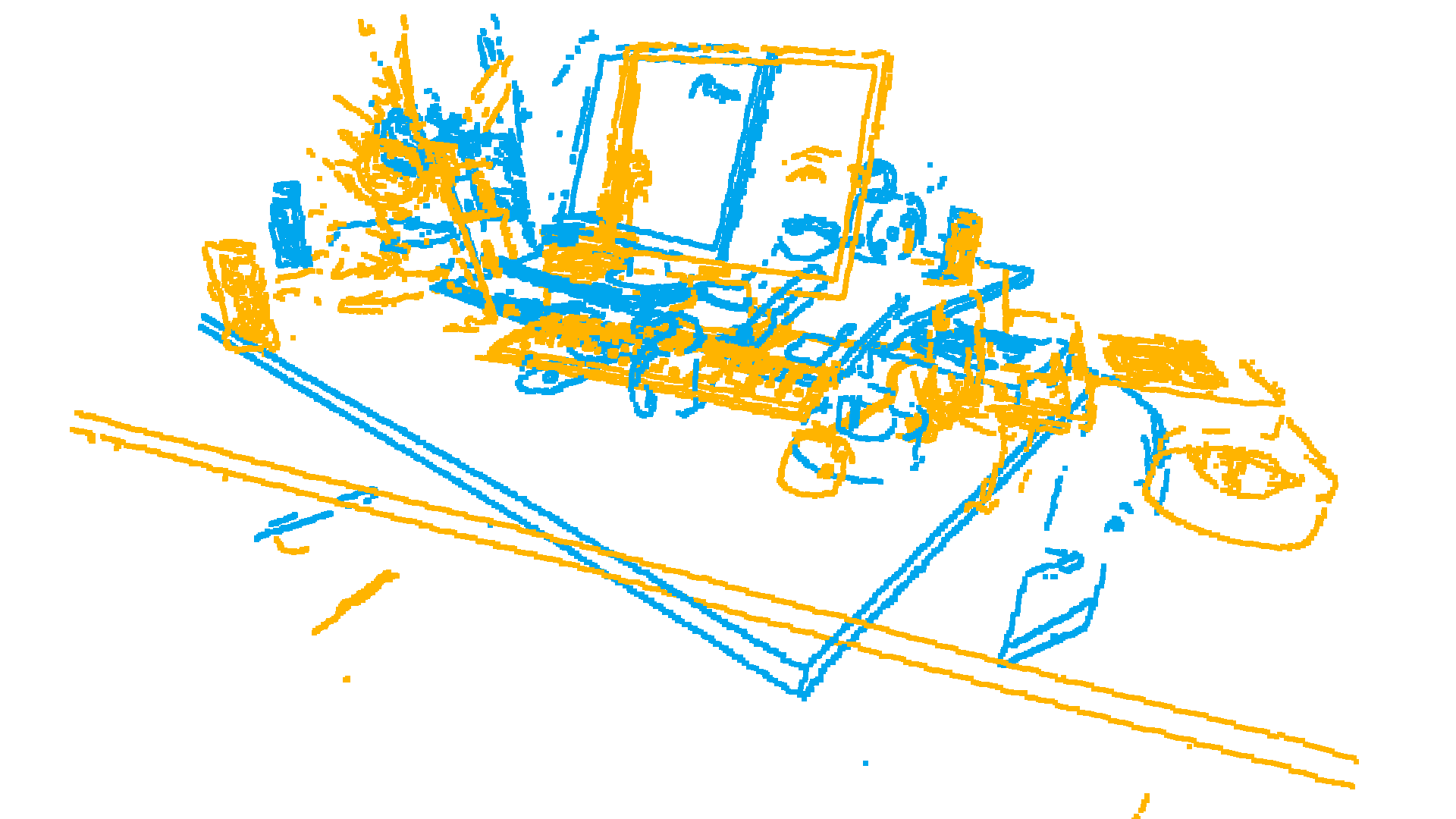}
        \put(50,-2){\color{black}\scriptsize\textbf{(a)}}
    \end{overpic}&
    \begin{overpic}[width=.5\columnwidth]{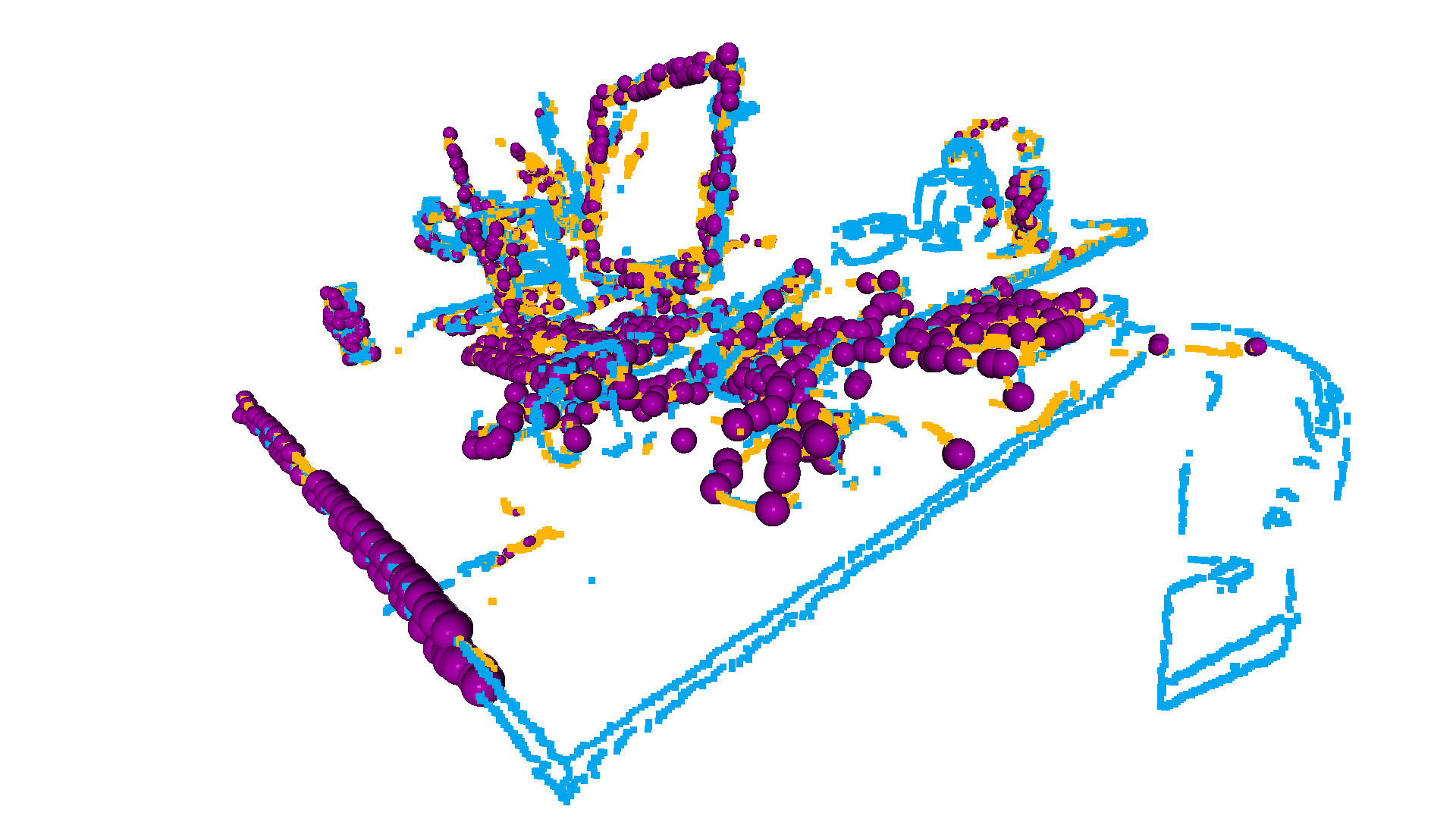}
        \put(50,-2){\color{black}\scriptsize\textbf{(b)}}
    \end{overpic}\\
    \begin{overpic}[width=.5\columnwidth]{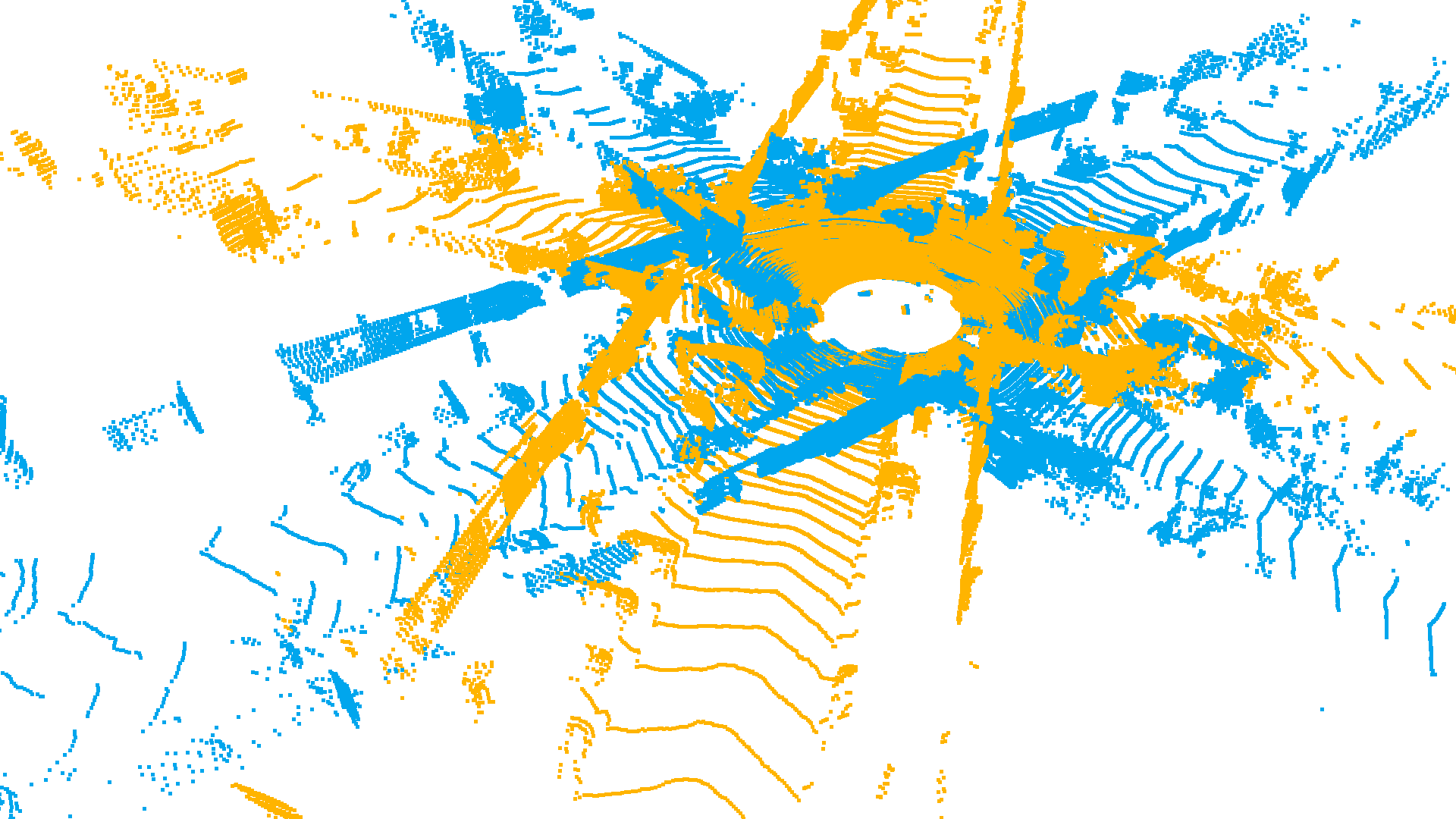}
        \put(50,-5){\color{black}\scriptsize\textbf{(c)}}
    \end{overpic}&
    \begin{overpic}[width=.5\columnwidth]{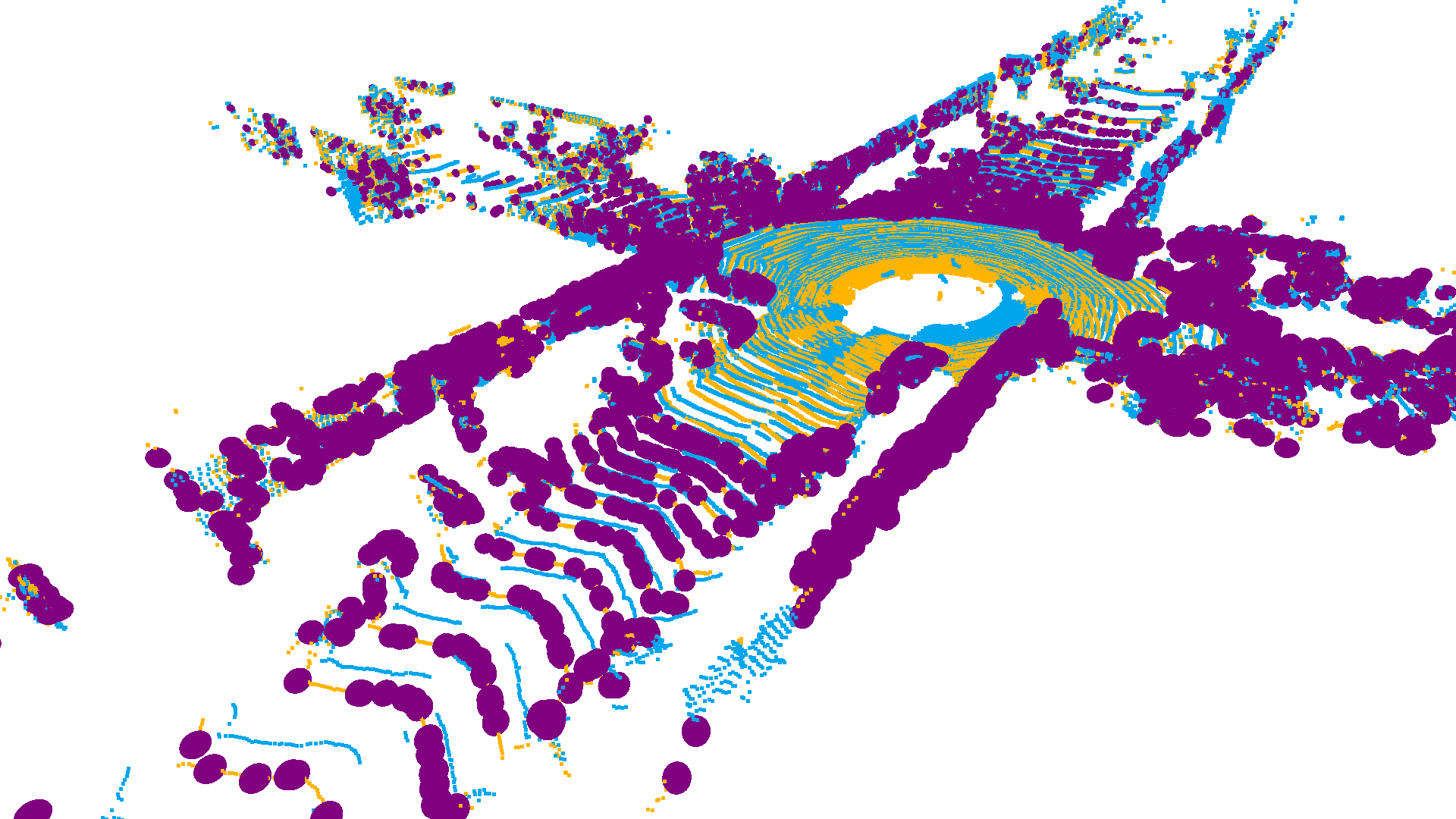}
        \put(50,-5){\color{black}\scriptsize\textbf{(d)}}
    \end{overpic}\\
  \end{tabular}
\end{center}
%\vspace{-.1cm}
\caption{Two examples of estimated overlap regions in the case of loops that show point clouds that are (a,b) reconstructed using RGBD SLAM \cite{Schenk2019} and (c,d) captured with LiDAR.
The pairs in the first column are displayed in their original camera reference frame, whereas in the second column they are displayed in a common reference frame after being registered with RANSAC using L3Ds. 
We highlight the overlap region estimated with our algorithm in purple.
The point clouds are taken (a,b) at frames 0 and 136 from the \texttt{fr2/desk} sequence of TUM-RGBD dataset \cite{Sturm2012} and (c,d) at frames 107 and 4532 from Sequence 00 of KITTI odometry dataset \cite{Geiger2012}.}
\label{fig:kitti_ron}
\end{figure}

Fig.~\ref{fig:kitti_ron} contains two examples of loops, and illustrates the estimated overlap regions of point clouds reconstructed using RGBD SLAM \cite{Schenk2019} and captured with LiDAR.
Although the L3Ds that we use are trained on a different domain than that of the examples, we can observe how our approach can successfully determine mutually-nearest neighbour points in the overlap region (purple) and how the estimated transformation can be used to register each point cloud pair.
In (d) we can observe that there are no mutually-nearest neighbours in the centre of the point clouds, this is due do the flat regions that carry little geometric information and the relative descriptors lack distinctiveness, so they cannot be matched reliably.

% ********************************
% ********************************

%%%%%%%%%%%%%%%%%%%%%%%%%%%%%%%%%%%%%%%%%%%%%%%%%%%%%%%%%%%%%%%%%%%%
%%%%%%%%%%%%%%%%%%%%%%%%%%%%%%%%%%%%%%%%%%%%%%%%%%%%%%%%%%%%%%%%%%%%
%%%%%%%%%%%%%%%%%%%%%%%%%%%%%%%%%%%%%%%%%%%%%%%%%%%%%%%%%%%%%%%%%%%%
\section{Experiments}
\label{sec:exp}

We evaluate our approach with two experiments using LiDAR SLAM and visual RGBD SLAM to show its general applicability.
Firstly, we compare our approach against recent LiDAR-based loop closure approaches, namely OverlapNet \cite{chen2021overlapnet} and LiDAR Iris \cite{IRIS2021}, using the LiDAR dataset KITTI odometry \cite{Geiger2012}. 
Secondly, we embed our loop closure method in a recent visual SLAM system, and test it on a real RGBD dataset, TUM-RGBD \cite{Sturm2012}, and on a synthetic RGBD dataset, ICL \cite{Saeedi2019}. 
We choose the edge-based visual SLAM method, namely RESLAM \cite{Schenk2019}, as the point clouds produced by the mapping are different from those of KITTI. 

\noindent\textbf{Local 3D deep descriptors.} We use DIP descriptors \cite{Poiesi2021} as they can achieve the best performance in terms of generalisation for point cloud registration applications (see Sec.~\ref{sec:descriptor_analysis} for more details), thus no need for retraining them when employed in new domains.
DIP descriptors are processed from an input point cloud in two main steps.
First, patches are extracted from the point cloud and for each patch a local reference frame is estimated to canonicalise the patch and thus to make the descriptor rotation-invariant.
Second, a deep neural network encodes the canonicalised patch into a 32-dimensional descriptor.
The deep network employs a PointNet backbone \cite{Qi2017a} that uses a Transformation Network (TNet) applied to the input points to improve the canonicalisation of the first step in the case of inaccurately estimated local reference frames.
The PointNet backbone is composed of a series of multilayer perceptrons that augment the input channels as 3$\rightarrow$256$\rightarrow$512$\rightarrow$1024, followed by max pooling to allow permutation invariance of the input points, which is then followed by another series of multilayer perceptrons to produce the output as 1024$\rightarrow$512$\rightarrow$256$\rightarrow$32.
TNet uses an equivalent architecture to this one, but produces a 3$\times$3 transformation as output.
These deep networks are trained on the 3DMatch dataset \cite{Zeng2017} through a Siamese approach using contrastive learning.
Please refer to \cite{Poiesi2021} for more details.
We use the same descriptor parameters as in \cite{Poiesi2021}, except for the radius of the patch. 
We use 2.5m as the patch radius for KITTI, and 0.2m for TUM RGBD and ICL.

%%%%%%%%%%%%%%%%%%%%%%%%%%%%%%%%%%%%%%%%%%%%%%%%%%%%%%%%%%%%%%%%%%%%
%%%%%%%%%%%%%%%%%%%%%%%%%%%%%%%%%%%%%%%%%%%%%%%%%%%%%%%%%%%%%%%%%%%%
\subsection{Loop closure detection for LiDAR SLAM}\label{sec:exp:lidar}

We compare our approach against two recent LiDAR-based methods \cite{chen2021overlapnet,IRIS2021} for loop closure detection.

\emph{OverlapNet} \cite{chen2021overlapnet} estimates the overlap with a deep neural network with the inputs extracted from the LiDAR scans containing both the geometric and semantic information. \emph{OverlapNet} defines the overlap between a scan pair as the ratio of the projected range pixels within 1m over all valid range pixels, where the candidate scan pairs are projected into range images with a common coordinate system.
For fair evaluation, we reproduce the results using the provided model with inputs reflecting only geometrical information, i.e., depths, normals and intensities.

\emph{LiDAR Iris} \cite{IRIS2021} uses a global descriptor for LiDAR scans. 
Each scan is converted into a binary signature image using a series of LoG-Gabor filtering and thresholding operations.
LiDAR Iris defines the overlap as the Hamming distance of a pair of the binary signature images. 
We reproduce the results using the provided code to generate the signature images and compute the Hamming distance.

Within this comparative evaluation, we also assess three variants of our L3D-based loop detection approach, namely 
\emph{L3D-based Overlap} that quantifies the overlap following the ground-truth overlap computation as in OverlapNet \cite{chen2021overlapnet}, \emph{L3D-based MNN} that quantifies the overlap only as the ratio of MNN (Eq.~\ref{eq:mnn}), and \emph{L3D-based RON} that quantifies the overlap based on the proposed RON (Eq.~\ref{eq:ron}).

%%%%%%%%%%%%%%%%%%%%%%%%%%%%%%%%%%%%%%%%%%%%%%%%%%%%%%%%%%%%%%%%%%%%
\noindent\textbf{Evaluation protocol.}
We follow the evaluation protocol proposed in OverlapNet \cite{chen2021overlapnet}, where a set of criteria are used for the detection of loops given a sequence.
We use Sequence 00 of KITTI odometry for the test.
As the vehicle moves, each current scan serves as a query.
We adopt two evaluation settings. 
\textit{Setting 1}: for each query, the latest 100 scans are excluded from the candidate loops in order to avoid detecting a loop closure in the most recent scans.
Like most of the SLAM systems, the candidate loop scans are constrained within the 3$\sigma$ area around the current pose estimate. 
We use the same series of pose uncertainty as in \cite{chen2021overlapnet} for the $\sigma$ estimation.
\textit{Setting 2}: we perform experiment where we lift the search constraint of the 3$\sigma$ area and the exclusion of the latest 100 scans. 
For each frame, we perform the search against all the previously seen frames to find the loop. 
In order to reduce the computation load in this setting, we first downsample the sequence by a factor of 20, forming a total of 25878 pairs, 51 of which are positive loop closures. 
Only the candidate loop scans with the highest estimated overlap are considered for the evaluation of these settings.
A true positive occurs if the ground-truth overlap is larger than 30\%.

% ********************************
\begin{figure}[t]
\centering
\includegraphics[width=0.45\textwidth]{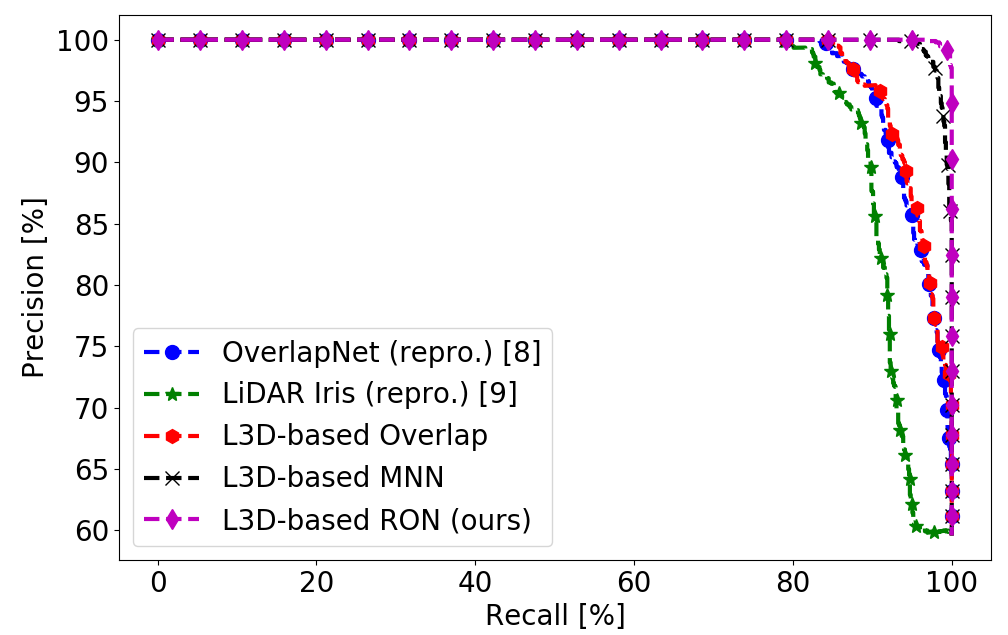}
\vspace{-.3cm}
\caption{The precision and recall curve of loop closure detection under Setting 1 with positional prior as a search  constraint.
We compare variants of our approach based on local 3D deep descriptors (L3D) and the state-of-the-art methods, i.e.~OverlapNet \cite{chen2021overlapnet} and LiDAR Iris \cite{IRIS2021}.
}
\label{fig:exp:lidarpr}
\end{figure}
% ********************************

%%%%%%%%%%%%%%%%%%%%%%%%%%%%%%%%%%%%%%%%%%%%%%%%%%%%%%%%%%%%%%%%%%%%
\noindent\textbf{Discussion.}
Fig.~\ref{fig:exp:lidarpr} illustrate the Precision-Recall (PR) curves under the Setting 1.
Note that no training on KITTI was performed with our L3D-based methods, while \emph{OverlapNet} is trained with the KITTI odometry using Sequences 03-10, and \emph{LiDAR Iris} is particularly designed for matching LiDAR scans.
Generalisation to unseen scenarios (sensors, environments) is a desired property for the deployment of robots in real-world applications.
\emph{L3D-based Overlap} produces a similar PR curve to OverlapNet, indicating that the relative transformations computed by RANSAC using the local 3D descriptors can achieve comparable performances to global loop closure detection methods.
Note that here we estimate the 6DoF transformation, while OverlapNet estimates the 1DoF transformation.
As shown by \emph{L3D-based MNN} curve, when we use the ratio of MNN in the descriptor space as the overlap measure, the PR curve further improves.
\emph{L3D-based RON} achieves the best PR curve, which justifies that the proposed overlap measure is more effective and reliable. To achieve a high RON value, it requires the points to be not only MNN in the descriptors space but also spatially close to each other after the estimated transformation.

\begin{figure}[t]
\centering
\includegraphics[width=0.42\textwidth]{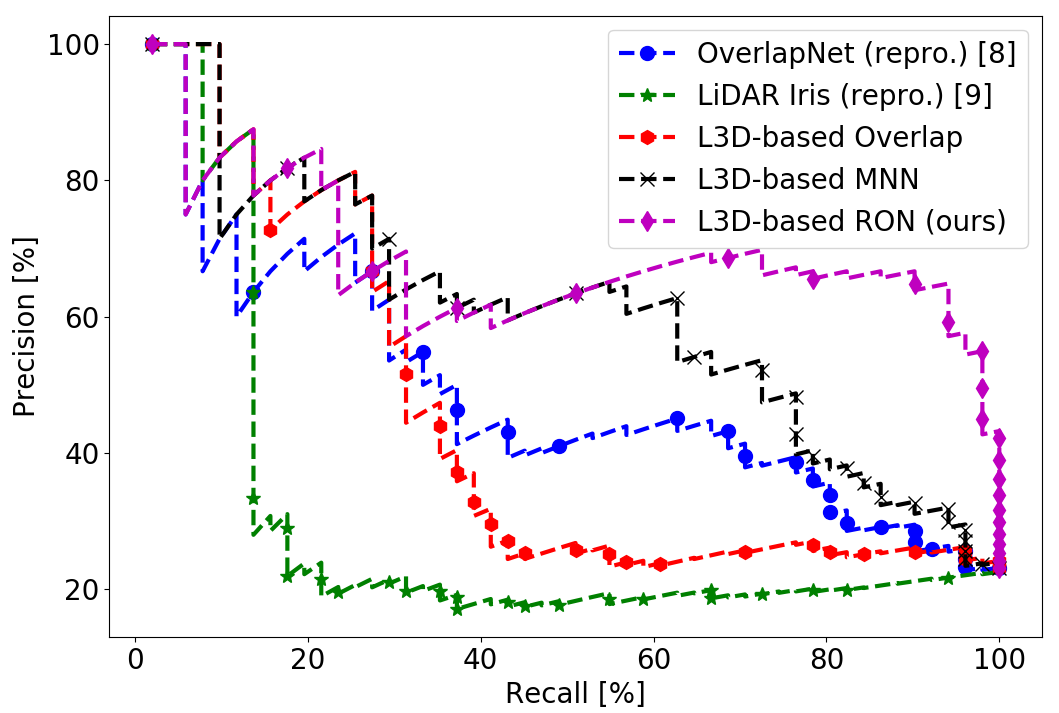}
\vspace{-.3cm}
\caption{The precision and recall curve of loop closure detection under Setting 2 without positional prior as a search  constraint.
We compare variants of our approach based on local 3D deep descriptors (L3D) against the state-of-the-art methods, i.e.~OverlapNet \cite{chen2021overlapnet} and LiDAR Iris \cite{IRIS2021}.}
\label{fig:exp:lidarpr_new}
\end{figure}

Fig.~\ref{fig:exp:lidarpr_new} reports the Precision-Recall (PR) curves under the Setting 2, where each frame is queried amongst all the previously seen frames to find the loop. 
Without the positional prior as a search constraint, all methods achieve a lower Precision values along the PR curves due to the large amount of False Positives as the threshold increases. Precision is generally low for all the methods because querying a point cloud against all the others at previous time steps leads to a higher likelihood of producing false positive loop detections due to similar neighbourhoods along the trajectory.
In terms of the Area Under Curve (AUC) performance, OverlapNet is better than LiDAR IRIS, while our proposed method \emph{L3D-based RON} is still the besting-performing method amongst all the other ones under this more challenging setup.

% ********************************
% ********************************

%%%%%%%%%%%%%%%%%%%%%%%%%%%%%%%%%%%%%%%%%%%%%%%%%%%%%%%%%%%%%%%%%%%%
%%%%%%%%%%%%%%%%%%%%%%%%%%%%%%%%%%%%%%%%%%%%%%%%%%%%%%%%%%%%%%%%%%%%
\subsection{Loop Closure for Visual SLAM}\label{sec:result_loop}
For the evaluation of visual RGBD SLAM, we use nine sequences from TUM-RGBD and seven sequences from ICL. 
TUM-RGBD contains real scenes recorded with a Kinect at 30Hz and its ground truth is captured using a motion capture system working at 100Hz. 
ICL contains photo-realistic scenes rendered with a simulated Kinect at 20-30Hz, featuring real-world trajectories captured using a motion capture system working at 100Hz.

We use the recent RESLAM~\cite{Schenk2019} to serve as the overall SLAM system where we embed our proposed approach. The original RESLAM performs the selection of candidate keyframe pairs based on the similarity between the current and the previous keyframes using 2D visual cues, i.e. Fern. If none of the previous keyframes yields a sufficient similarity with the current one, the loop closure is terminated, and a global and compact representation of the new keyframe is added to a database. Otherwise a geometric-based assessment is further carried out using pose graph optimisation.

We use the original parameters of RESLAM and set $n=0.4$, $\tau_o=0.13$, $\tau_e=10$cm and $\tau_\rho=0.2$ (see notations in Sec.~\ref{sec:loop_det}).
We set the max number of RANSAC iterations to 500K and the termination policy to 1K iterations for an average inlier error of 3cm.
We report the results of RESLAM from their paper \cite{Schenk2019} and also include our reproduced results as the authors of \cite{Schenk2019} stated that different hardware may provide different results.

We use RESLAM [repro.1] and RESLAM [repro.2] to refer to the results reproduced by us using two configurations of RESLAM: we use the original parameters in the former, while we set the parameters to detect loop closures more frequently in the latter.
The evaluation of the latter was performed to show the behaviour of RESLAM with a number of detected loops comparable to those of our approach.
We also evaluate an ablated version of our approach, i.e.~only the module to estimate the transformation matrix between the keyframe and loop frame based on L3D, i.e. DIP \cite{Poiesi2021} in our experiments.
This ablation experiment is designed to show the benefit of using L3D for pose estimation after the loop closures detected with RESLAM's visual cues.
We name this version RESLAM [w.~L3D].

%%%%%%%%%%%%%%%%%%%%%%%%%%%%%%%%%%%%%%%%%%%%%%%%%%%%%%%%%%%%%%%%%%%%
\noindent \textbf{Evaluation metrics.}
We evaluate our approach by comparing the absolute errors between the estimated and the ground-truth trajectories, noted as ATE, by computing the root mean squared error (RMSE) of the translational component \cite{Sturm2012}.
Given that the rigid-body transformation $\mathbf{S}$ corresponding to the least-squares solution that maps the estimated pose of the camera $\mathbf{C}_t$ onto its ground-truth pose $\mathbf{G}_t$ at frame $t$, ATE can be computed as $\mathcal{E}_t = \mathbf{G}_t^{-1} \mathbf{S} \mathbf{C}_t$.
The RMSE is computed over all the frames of each sequence as
%+++++++++++++++++++++
\begin{equation}\label{eq:rmse}
    RMSE_\mathcal{E} = \sqrt{ \frac{1}{n} \sum_{t=1}^{t_\mathsf{end}} \lVert \mathsf{trans}(\mathcal{E}_t)\rVert^2 },
\end{equation}
%+++++++++++++++++++++
where $t_\mathsf{end}$ is the number of frames of a given sequence.

%+++++++++++++++++++++++++
\begin{table}[t]
    \tabcolsep 3pt
    \centering
    \caption{
    RMSE [cm] (Eq.~\ref{eq:rmse}) computed on the RGBD-TUM dataset.
    The number of detected loops are reported in the parentheses.}
    \vspace{-.2cm}
    \label{tab:quant_results}  
    \resizebox{1\columnwidth}{!}{%
    \begin{tabular}{lccccc}
        \toprule
        \multirow{2}{*}{seq.} & {RESLAM} & RESLAM & RESLAM & RESLAM & \multirow{2}{*}{OURS} \\
        & \cite{Schenk2019} & [repro.1] & [repro.2] & [w.~L3D] & \\
        \midrule
        \texttt{\footnotesize fr1/xyz} & \emph{1.1} & 2.3 (27) & 2.3 (27) & 2.3 (27) & 1.8 (27) \\
        \texttt{\footnotesize fr1/room} & - & 8.6 (3) & 9.6 (11) & 8.5 (3) & 6.4 (14) \\
        \texttt{\footnotesize fr1/plant} & - & 8.6 (1) & 6.7 (6) & 7.3 (3) & 7.0 (5) \\
        \texttt{\footnotesize fr1/desk} & - & 2.7 (5) & 2.7 (7) & 2.7 (5) & 2.7 (7) \\
        \texttt{\footnotesize fr1/rpy} & - & 2.9 (19) & 2.6 (25) & 2.9 (19) & 2.6 (25) \\
        \texttt{\footnotesize fr1/desk2} & \emph{4.8} & 6.3 (7) & 5.8 (15) & 3.8 (8) & 3.6 (17) \\
        \texttt{\footnotesize fr2/desk} & \emph{1.9} & 2.2 (24) & 128.8 (25) & 2.2 (24) & 2.2 (38)\\
        \texttt{\footnotesize fr2/xyz} & \emph{0.5} & 0.4 (14) & 0.4 (15) & 0.5 (16) & 0.5 (16) \\
        \texttt{\footnotesize fr3/office} & \emph{3.5} & 3.8 (8) & 24.0 (15) & 3.8 (10) & 3.4 (15) \\
        \midrule
        average & & 4.2 & 20.3 & 3.8 & \textbf{3.4} \\
        \bottomrule 
    \end{tabular}
    }
\end{table}
%+++++++++++++++++++++++++

%+++++++++++++++++++++++++
\begin{table}[t]
    \tabcolsep 5pt
    \centering
    \caption{
    RMSE [cm] (Eq.~\ref{eq:rmse}) computed on the ICL dataset.
    The number of detected loops are reported in the parentheses.}
    \vspace{-.2cm}
    \label{tab:quant_results1}
    \resizebox{1\columnwidth}{!}{%
    \begin{tabular}{lcccc}
        \toprule
        \multirow{2}{*}{seq.} & RESLAM & RESLAM & RESLAM & \multirow{2}{*}{OURS} \\
        & [repro.1] & [repro.2] & [w.~L3D] & \\
        \midrule
        \texttt{\footnotesize deer/walk} & 10.1 (15) & {30.7} (33) & 8.6 (26) & 6.0 (48) \\
        \texttt{\footnotesize deer/Mfast} & 1.0 (89) & {1.0} (92) & 1.0 (82) & 1.0 (106) \\
        \texttt{\footnotesize deer/Mslow} & 1.7 (110) & {3.6} (126) & 1.7 (112) & 1.6 (139) \\
         \texttt{\footnotesize deer/run} & 5.4 (4) & {19.1} (9) & 3.5 (7) & 2.6 (16) \\
         \texttt{\footnotesize diamond/walk} & 7.2 (9) & {19.3} (22) & 7.4 (11) & 3.8 (46) \\
         \texttt{\footnotesize diamond/Mfast} & 1.0 (87) & {0.9} (92) & 1.0 (89) & 0.9 (105) \\
         \texttt{\footnotesize diamond/run} & 14.4 (2) & {10.8} (7) & 8.6 (6) & 6.0 (10) \\
         \midrule
         average & 5.8 & 12.1 & 4.5 & \textbf{3.1} \\
        \bottomrule
    \end{tabular}
    }
\end{table}
%+++++++++++++++++++++++++

% ********************************
\begin{figure}[t!]
\begin{center}
  \begin{tabular}{@{}c@{\,\,\,\,\,}c@{}c}
  %%%%%%%%%%%%%%%%%%%%%%%%%%
  \adjustbox{raise=0cm}{\begin{overpic}[width=.43\columnwidth]{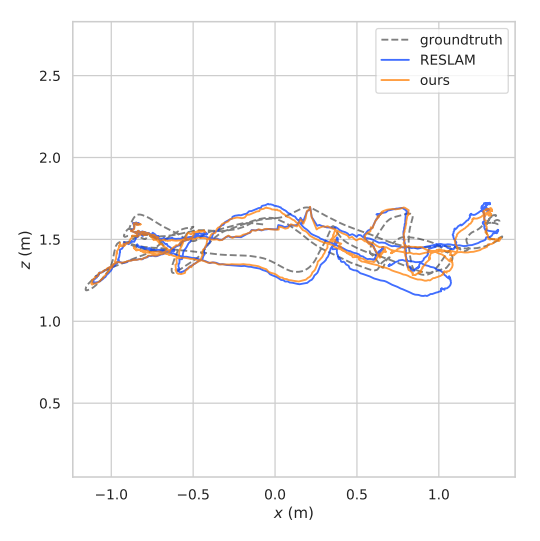}
      \put(55,43){
            \begin{tikzpicture}
            \draw[violet, very thick, opacity=0.65] (0,0) rectangle (.96,.62);
            \end{tikzpicture}
        }
        \put(6,-2){\color{black}\scriptsize\textbf{a) fr1/room}}
    \end{overpic}}&
    \adjustbox{raise=.97cm}{\begin{overpic}[width=.5\columnwidth, height=.3\columnwidth]{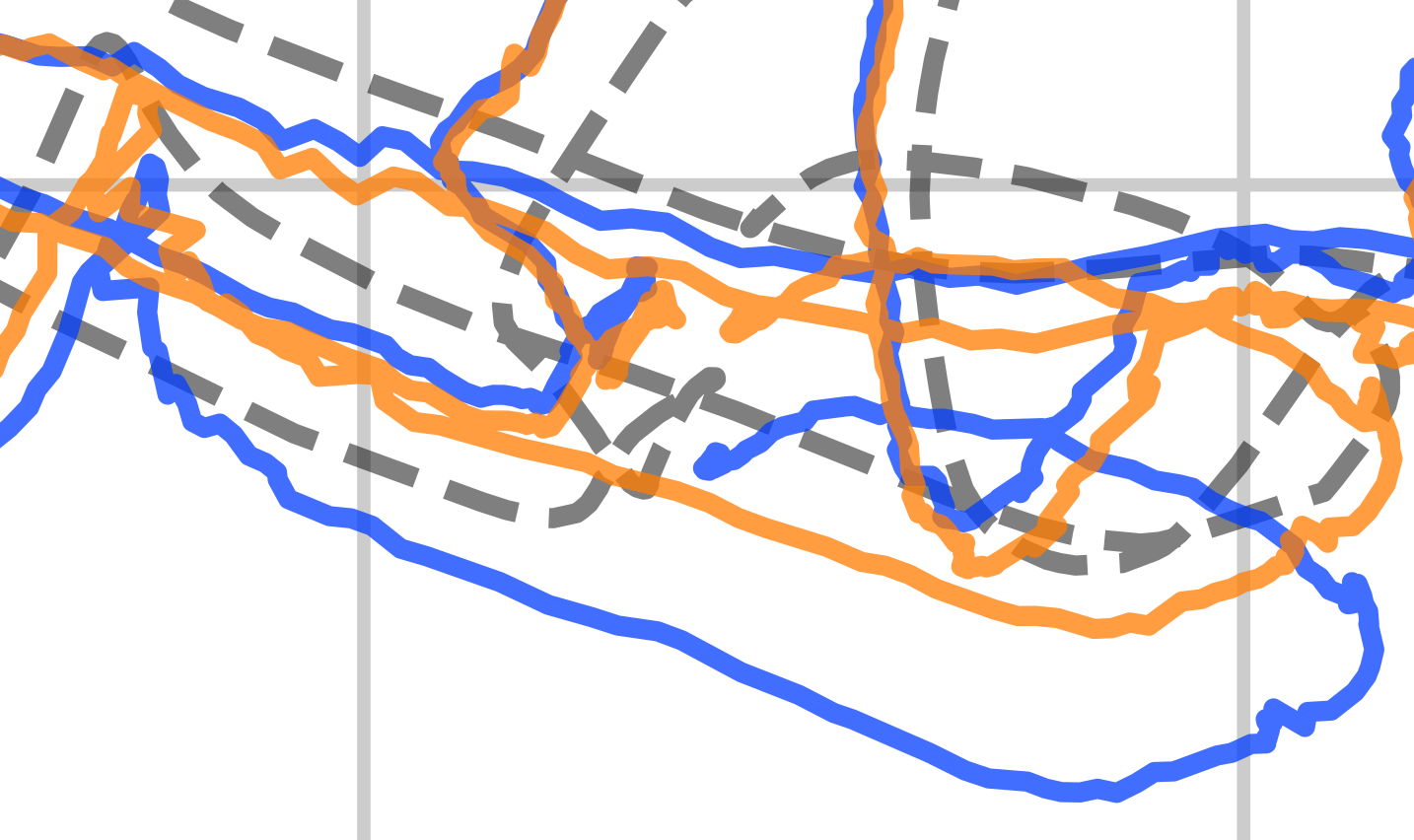}
        \put(-3,0){
                \begin{tikzpicture}
                \draw[violet, very thick, opacity=0.65] (0,0) rectangle (4.3,2.55);
                \end{tikzpicture}
                }
        \put(23,-7){\color{black}\scriptsize zoomed-in bounding box}
    \end{overpic}}&\\
    %%%%%%%%%%%%%%%%%%%%%%%%%%
    \adjustbox{raise=0cm}{\begin{overpic}[width=.43\columnwidth]{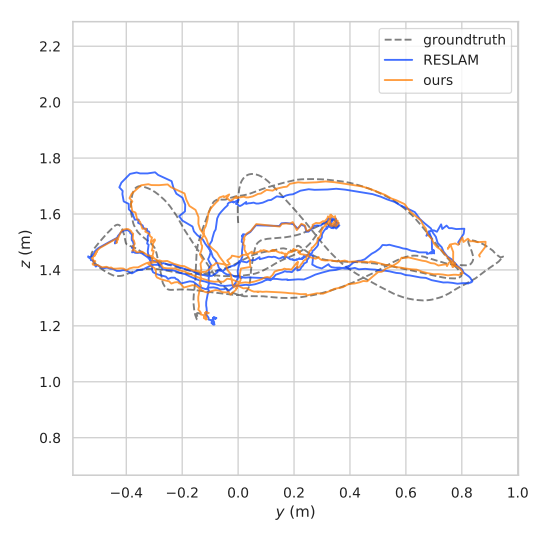}
      \put(42,55){
            \begin{tikzpicture}
            \draw[violet, very thick, opacity=0.65] (0,0) rectangle (.86,.51);
            \end{tikzpicture}
        }
        \put(6,-2){\color{black}\scriptsize\textbf{b) fr1/desk2}}
    \end{overpic}}&
    \adjustbox{raise=.97cm}{\begin{overpic}[width=.5\columnwidth, height=.3\columnwidth]{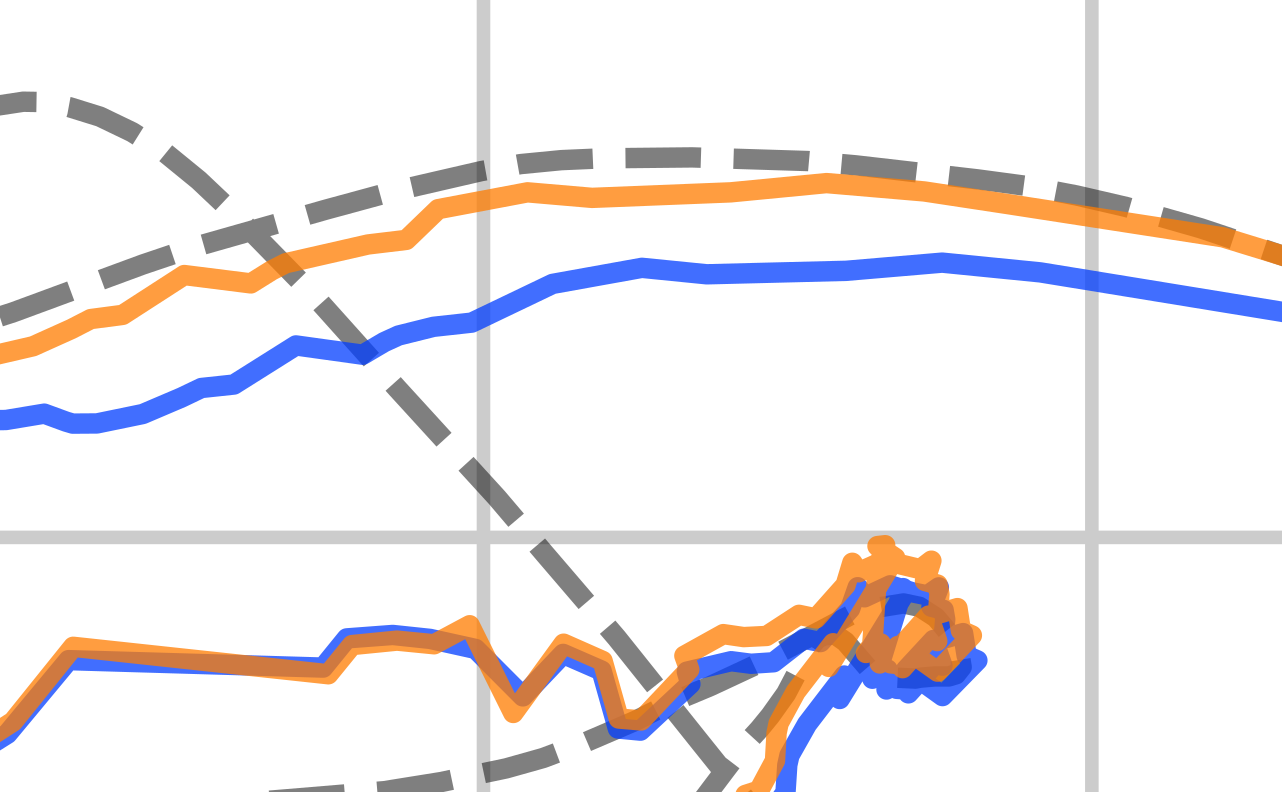}
        \put(-3,0){
                \begin{tikzpicture}
                \draw[violet, very thick, opacity=0.65] (0,0) rectangle (4.3,2.55);
                \end{tikzpicture}
                }
        \put(23,-7){\color{black}\scriptsize zoomed-in bounding box}
    \end{overpic}}&\\
    %%%%%%%%%%%%%%%%%%%%%%%%%%
    \adjustbox{raise=0cm}{\begin{overpic}[width=.43\columnwidth]{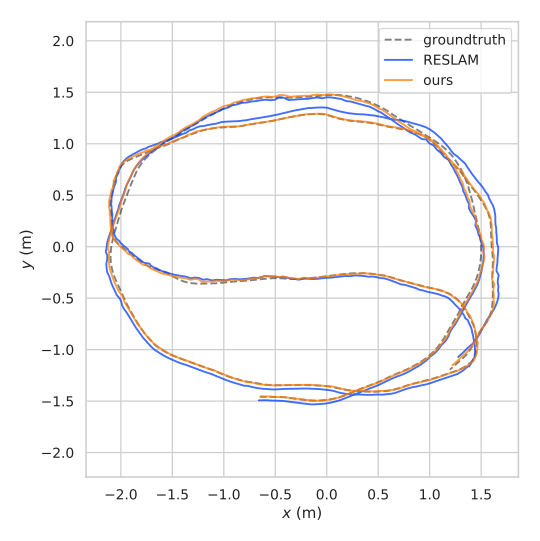}
      \put(44,71){
            \begin{tikzpicture}
            \draw[violet, very thick, opacity=0.65] (0,0) rectangle (.77,.52);
            \end{tikzpicture}
        }
        \put(6,-2){\color{black}\scriptsize\textbf{c) deer/run}}
    \end{overpic}}&
    \adjustbox{raise=.97cm}{\begin{overpic}[width=.5\columnwidth, height=.3\columnwidth]{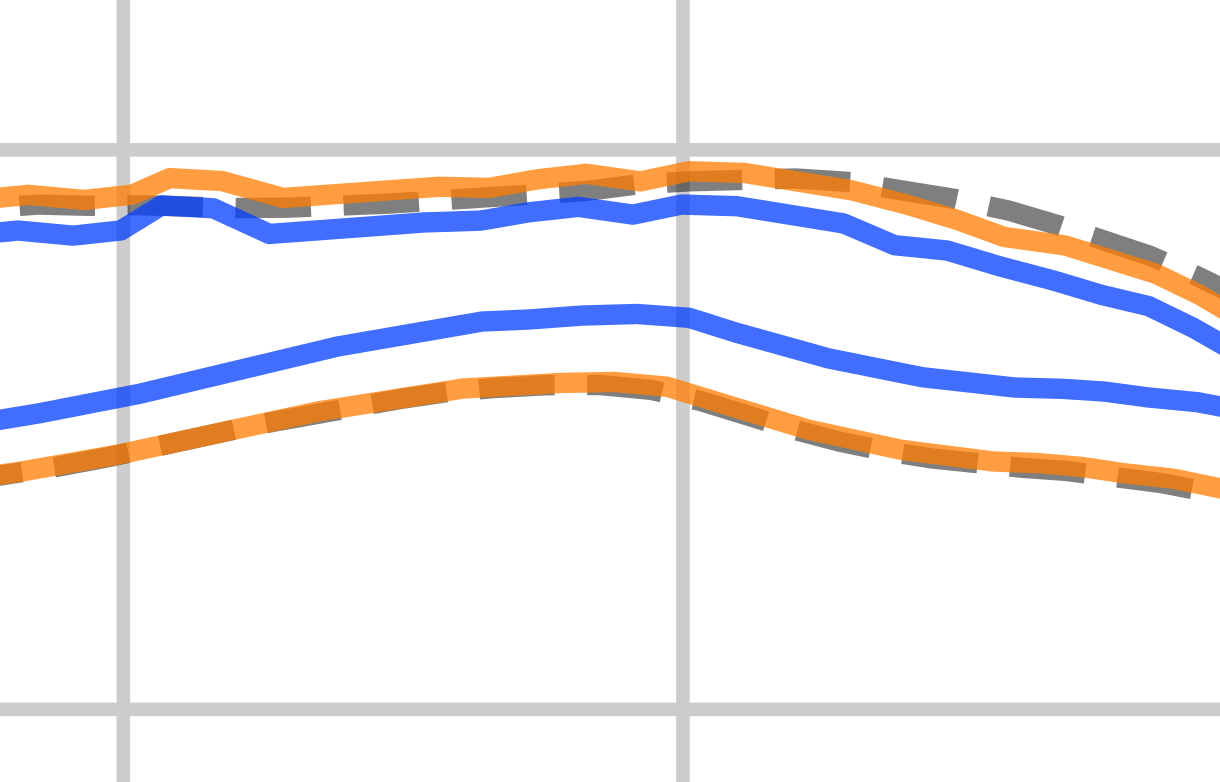}
        \put(-3,0){
                \begin{tikzpicture}
                \draw[violet, very thick, opacity=0.65] (0,0) rectangle (4.3,2.55);
                \end{tikzpicture}
                }
        \put(23,-7){\color{black}\scriptsize zoomed-in bounding box}
    \end{overpic}}&\\
  \end{tabular}
\end{center}
\vspace{-.3cm}
\caption{Trajectories obtained with the original RESLAM's loop closure approach and with our approach on (a,b) TUM-RGBD and (c) ICL sequences, projected on the $xz$, $yz$ and $xy$ planes, respectively.
Purple bounding boxes show examples where our approach improves localisation.
The right-hand side column is the zoomed-in version of the purple bounding box.}
\label{fig:qualitative_res_traj}
\end{figure}
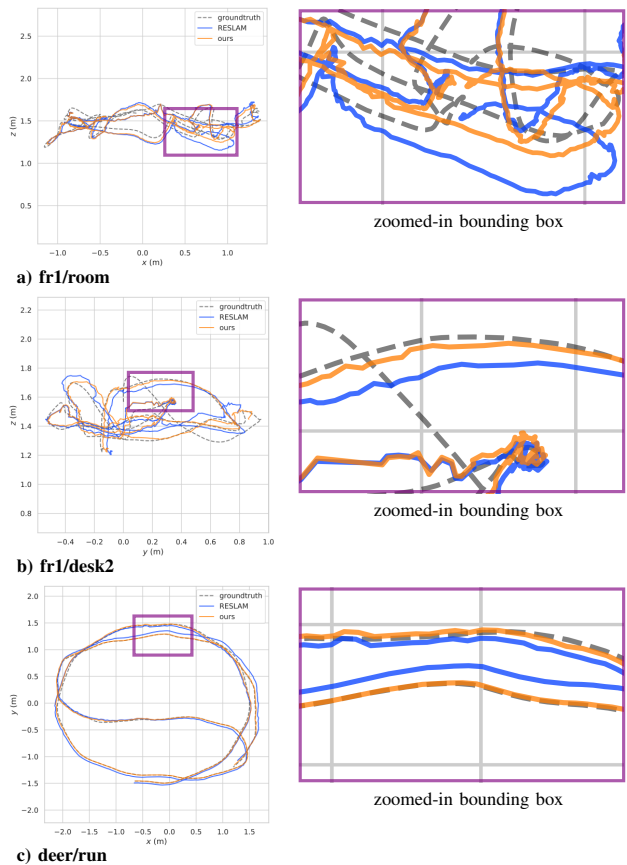
% ********************************

% ********************************
\begin{figure}[t!]
%\vspace*{3mm}
\begin{center}
  \begin{tabular}{@{}c@{}c@{}c}
  %%%%%%%%%%%%%%%%%%%%%%%%%%
    \begin{overpic}[width=.33\columnwidth]{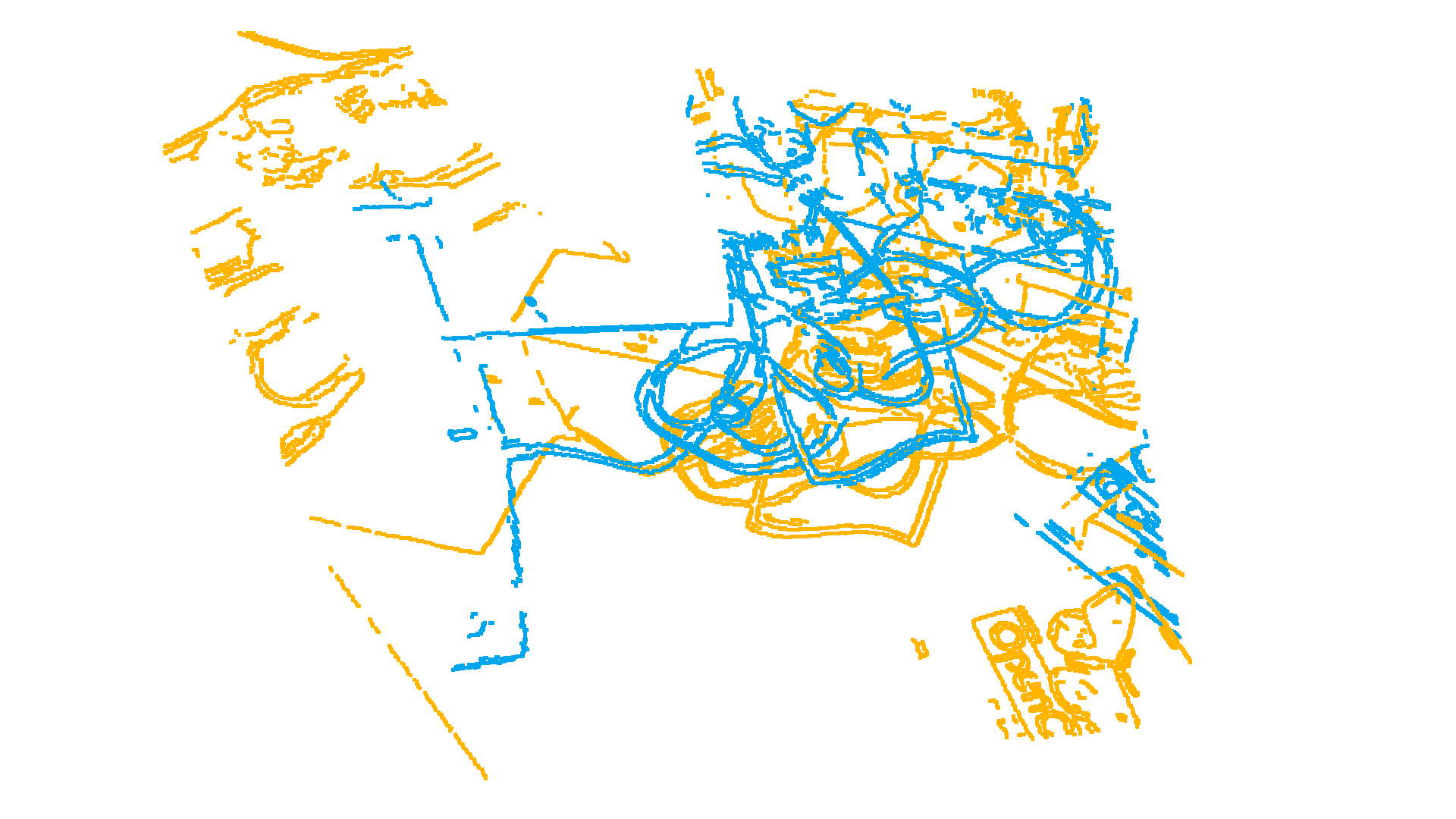}
        \put(6,-1){\color{black}\scriptsize\textbf{a) fr1/desk}}
        \put(30,60){\color{black}\scriptsize\textbf{original keyframes}}
    \end{overpic}&
    \begin{overpic}[width=.33\columnwidth]{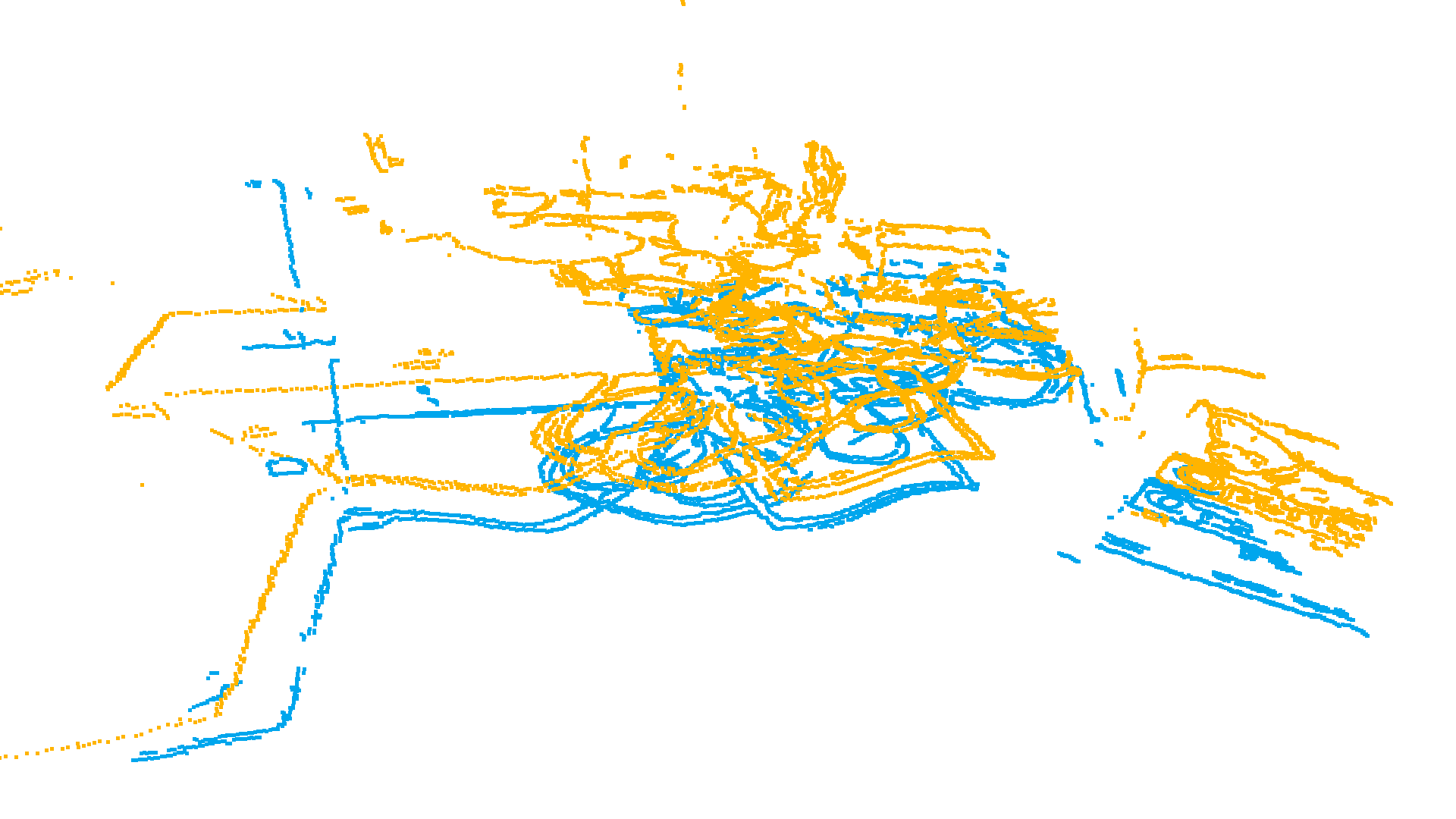}
        \put(30,60){\color{black}\scriptsize\textbf{RESLAM \cite{Schenk2019}}}
    \end{overpic}&
    \begin{overpic}[width=.33\columnwidth]{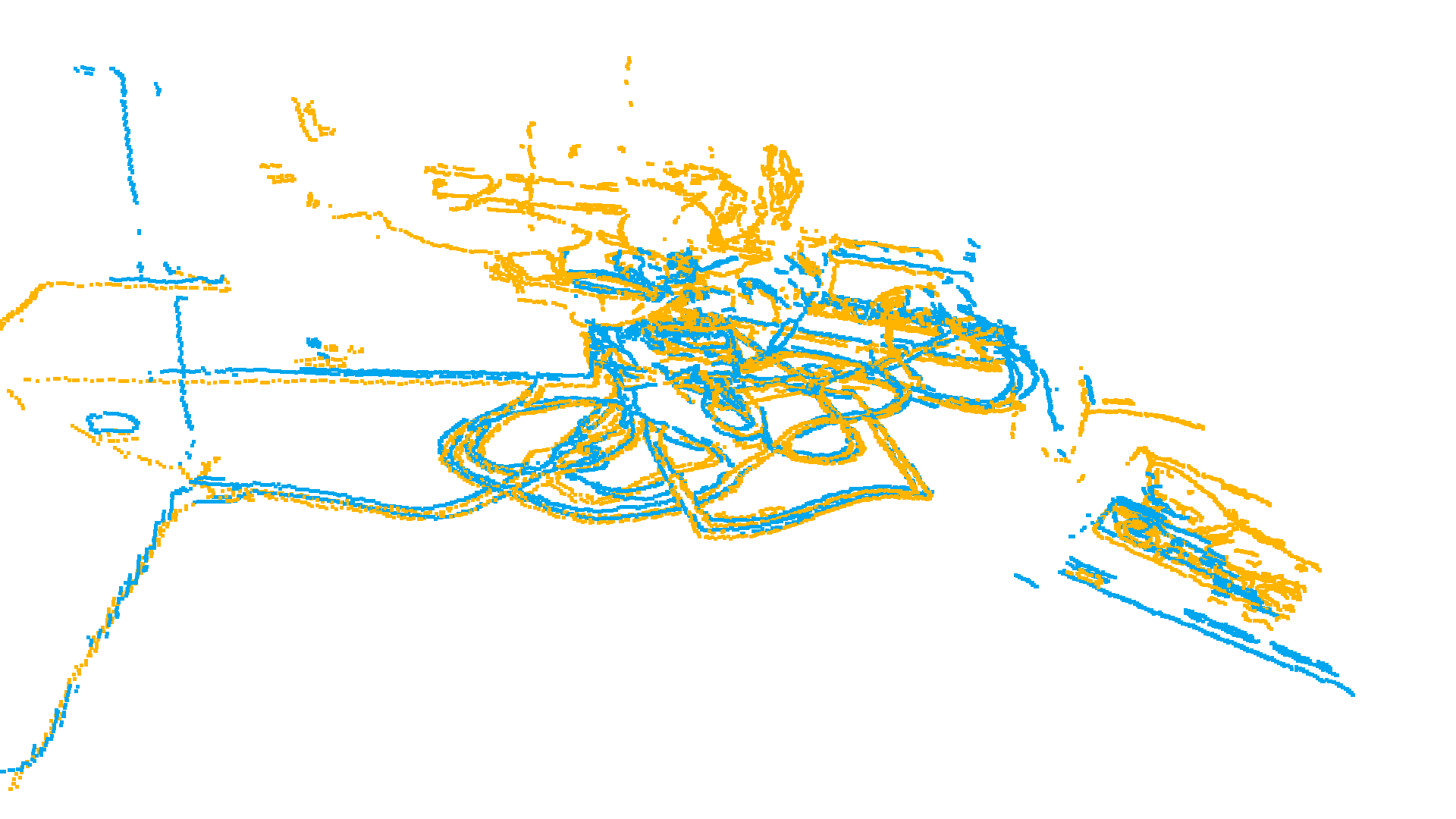}
        \put(35,60){\color{black}\scriptsize\textbf{ours}}
    \end{overpic}\\
    %%%%%%%%%%%%%%%%%%%%%%%%%%
    \begin{overpic}[width=.33\columnwidth]{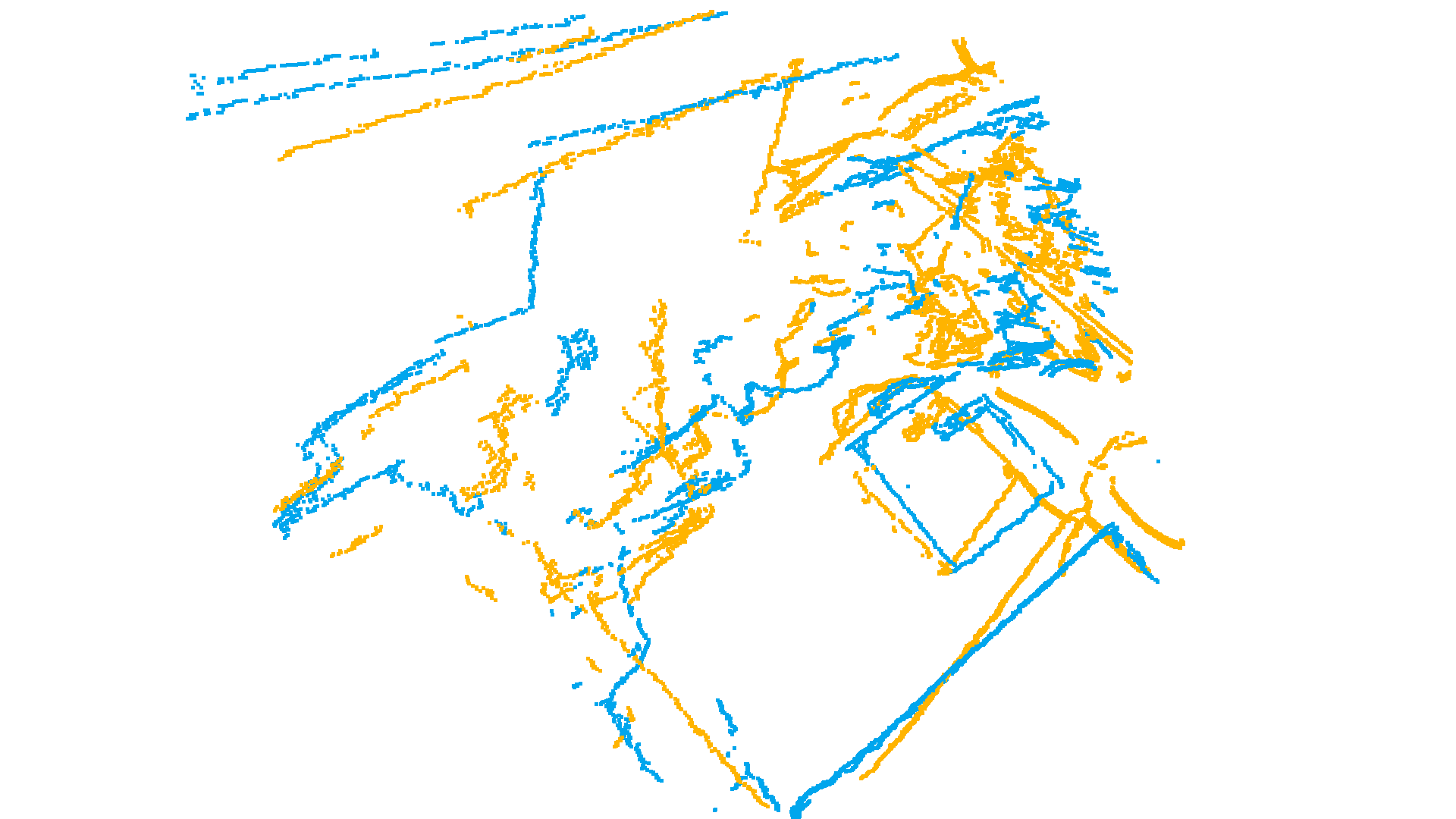}
        \put(6,-1){\color{black}\scriptsize\textbf{b) fr1/room}}
    \end{overpic}&
    \begin{overpic}[width=.33\columnwidth]{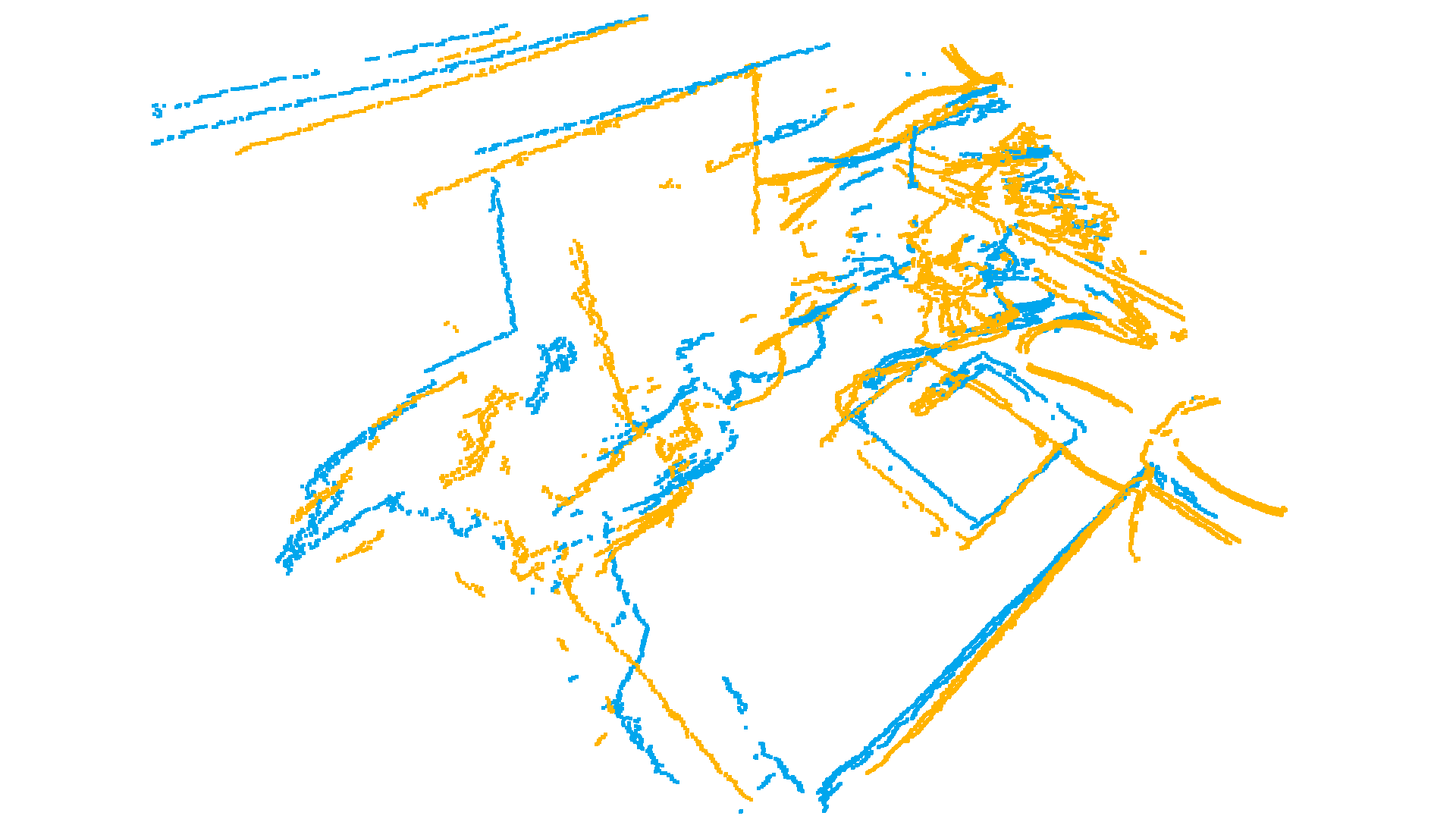}
    \end{overpic}&
    \begin{overpic}[width=.33\columnwidth]{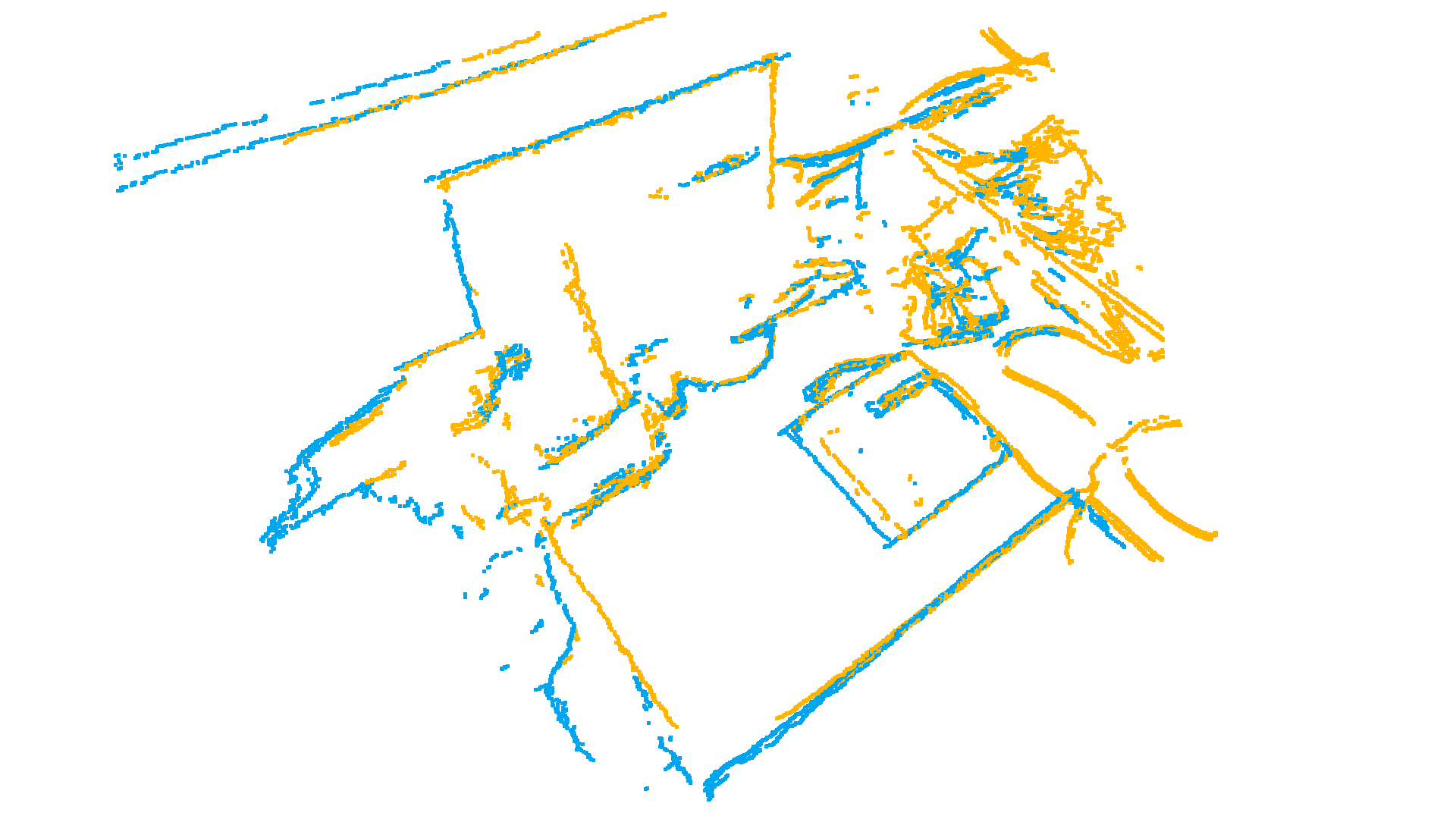}
    \end{overpic}\\
    %%%%%%%%%%%%%%%%%%%%%%%%%%
    \begin{overpic}[width=.33\columnwidth]{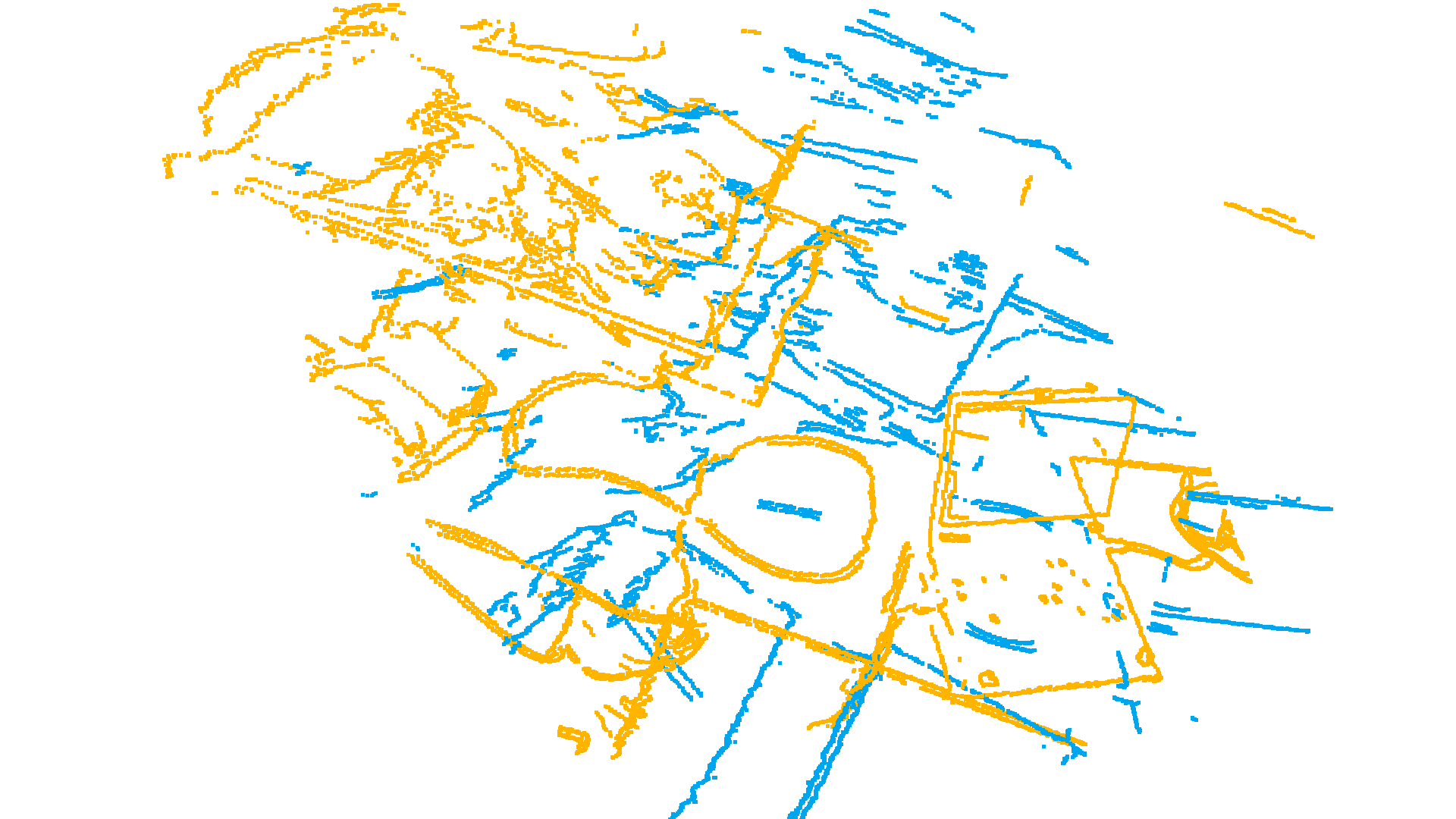}
        \put(6,-1){\color{black}\scriptsize\textbf{c) fr1/room}}
    \end{overpic}&
    \begin{overpic}[width=.33\columnwidth]{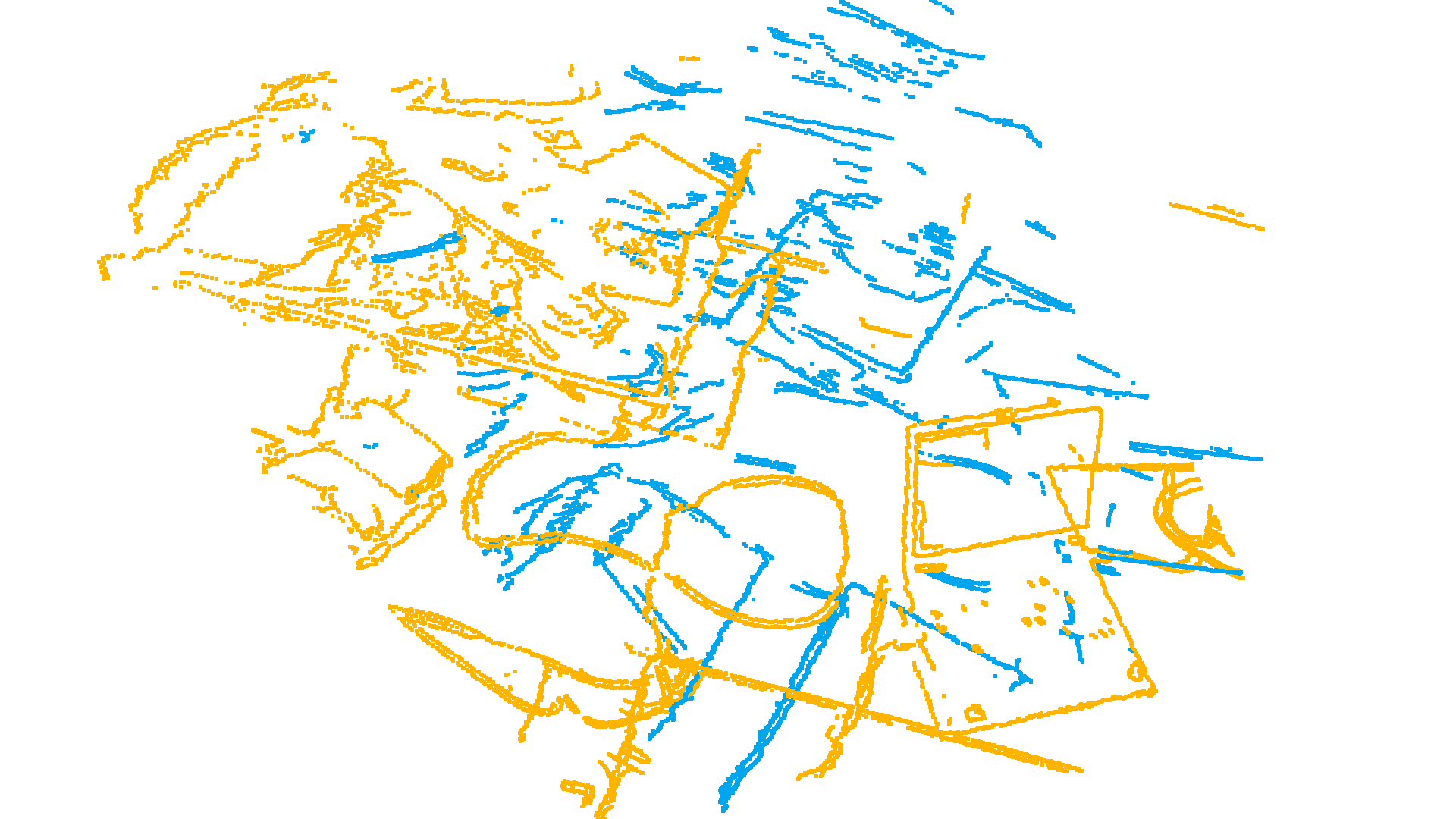}
    \end{overpic}&
    \begin{overpic}[width=.33\columnwidth]{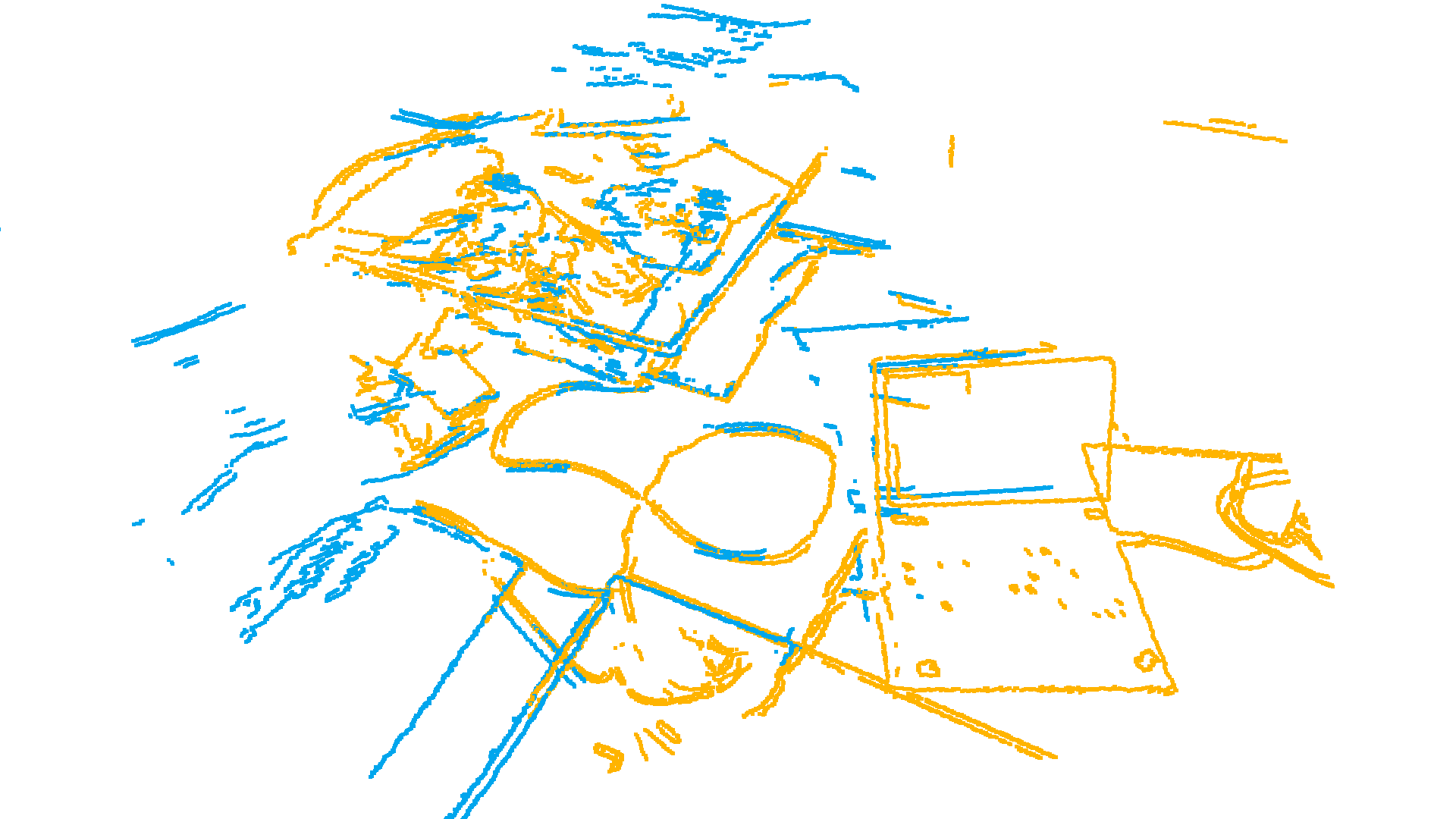}
    \end{overpic}\\
    %%%%%%%%%%%%%%%%%%%%%%%%%%
    \begin{overpic}[width=.33\columnwidth]{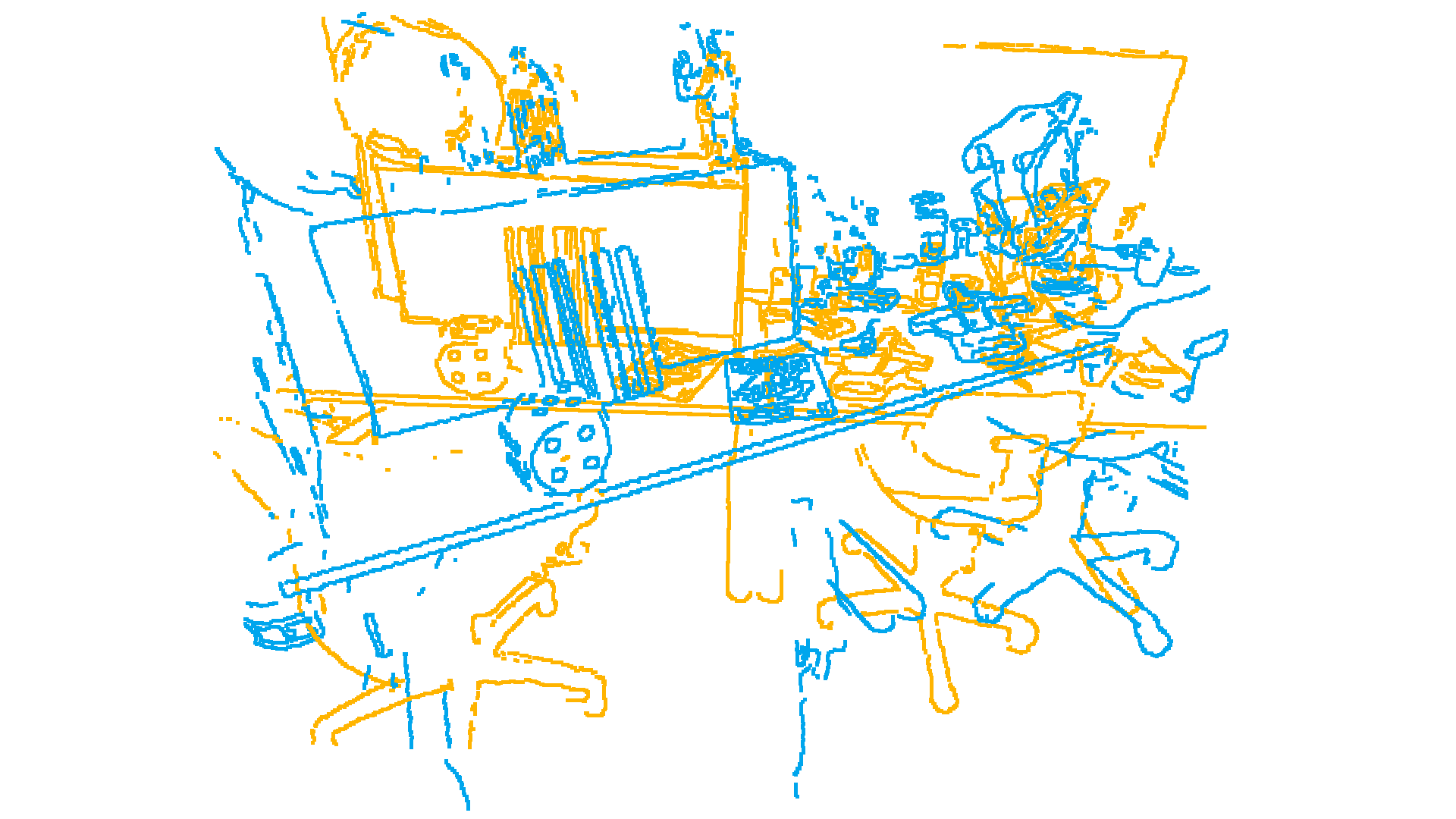}
        \put(6,-4){\color{black}\scriptsize\textbf{d) fr3/office}}
    \end{overpic}&
    \begin{overpic}[width=.33\columnwidth]{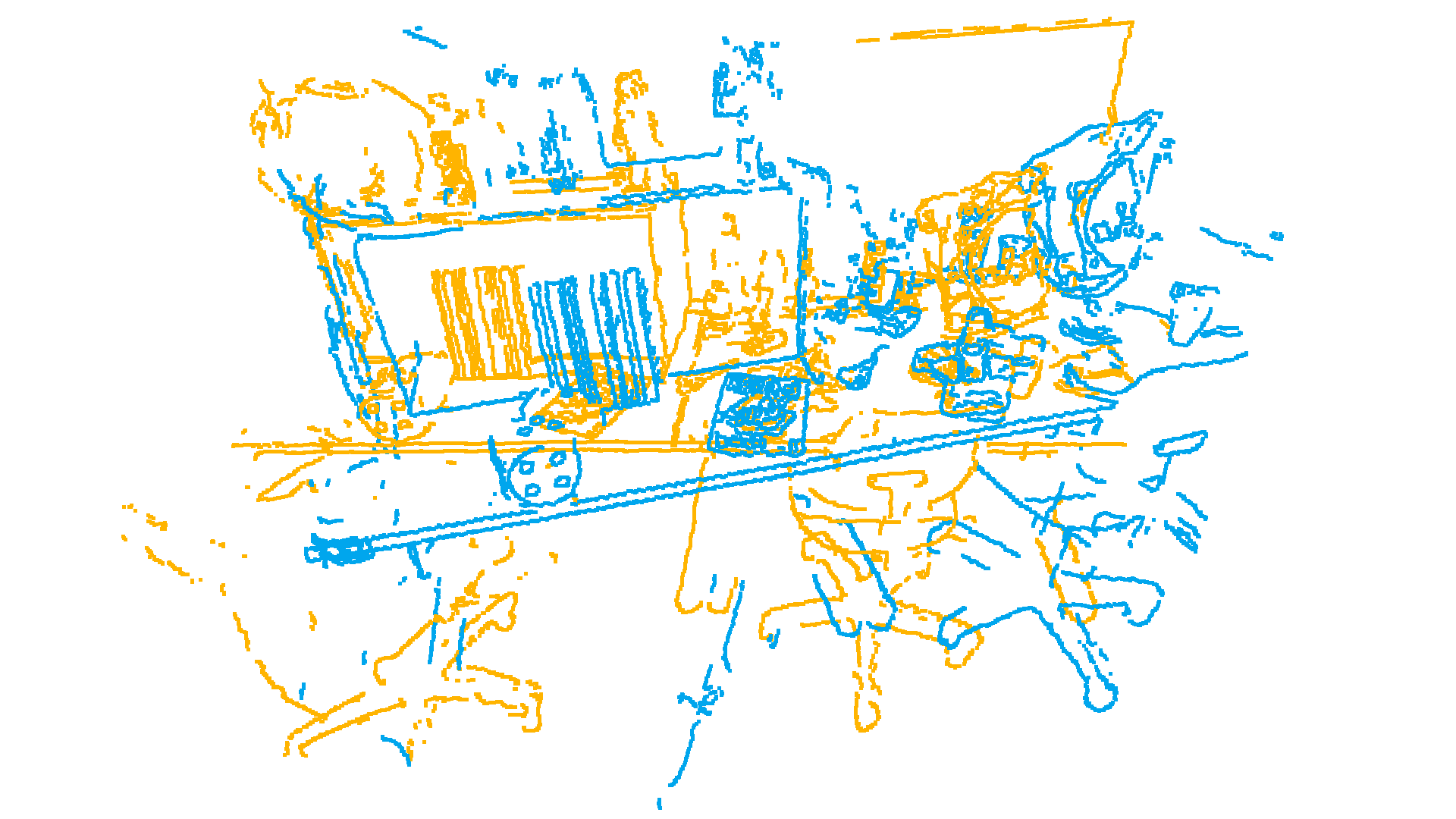}
    \end{overpic}&
    \begin{overpic}[width=.33\columnwidth]{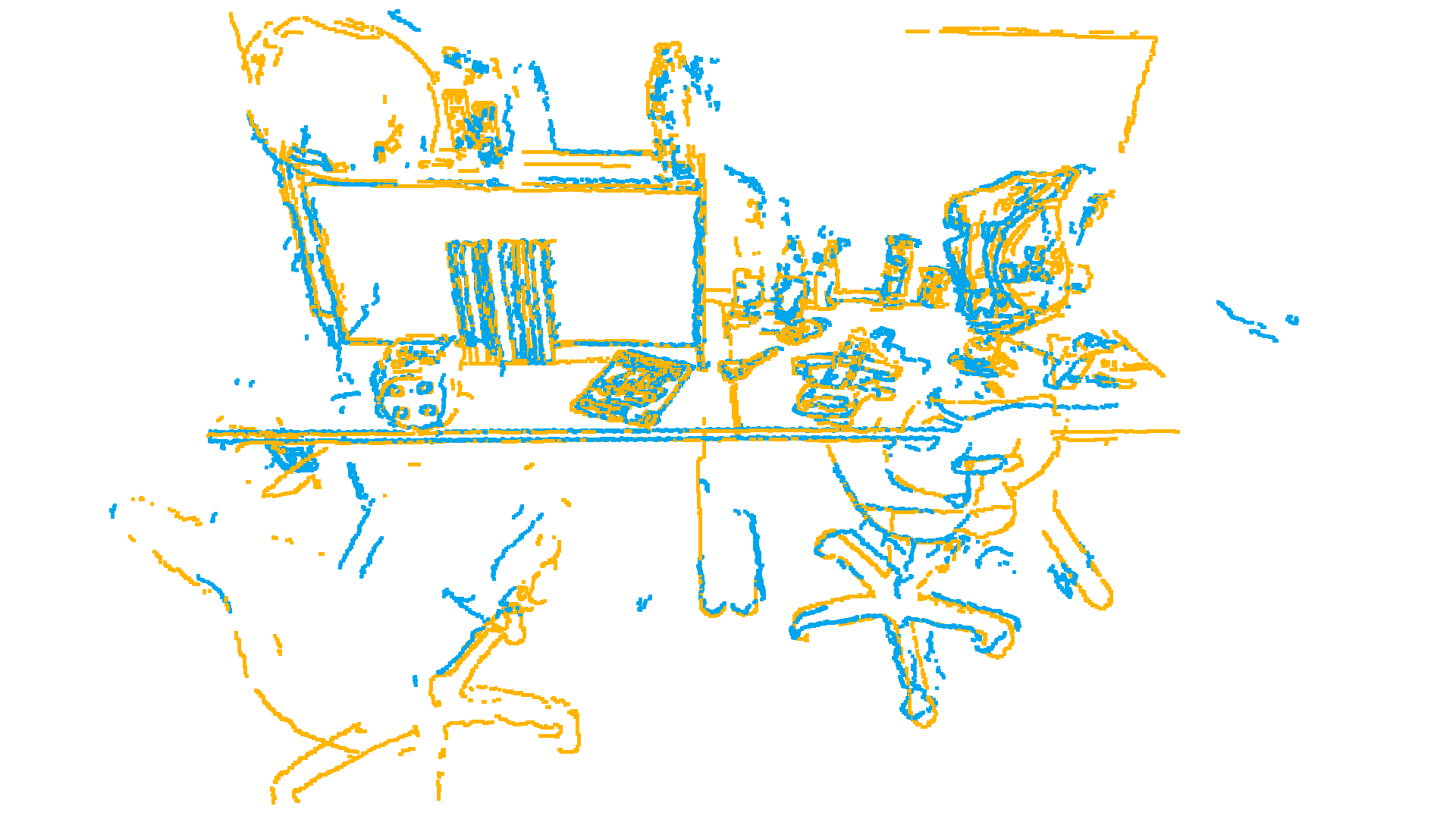}
    \end{overpic}\\
  \end{tabular}
\end{center}
\vspace{-.2cm}
\caption{Edge point cloud registration results using the original RESLAM's loop closure approach and ours.}
\label{fig:qualitative_res_pc_registration}
\end{figure}
% ********************************

% ********************************
\begin{figure}[t!]
\begin{center}
  \begin{tabular}{@{}c@{}c@{}c}
    \begin{overpic}[width=.33\columnwidth]{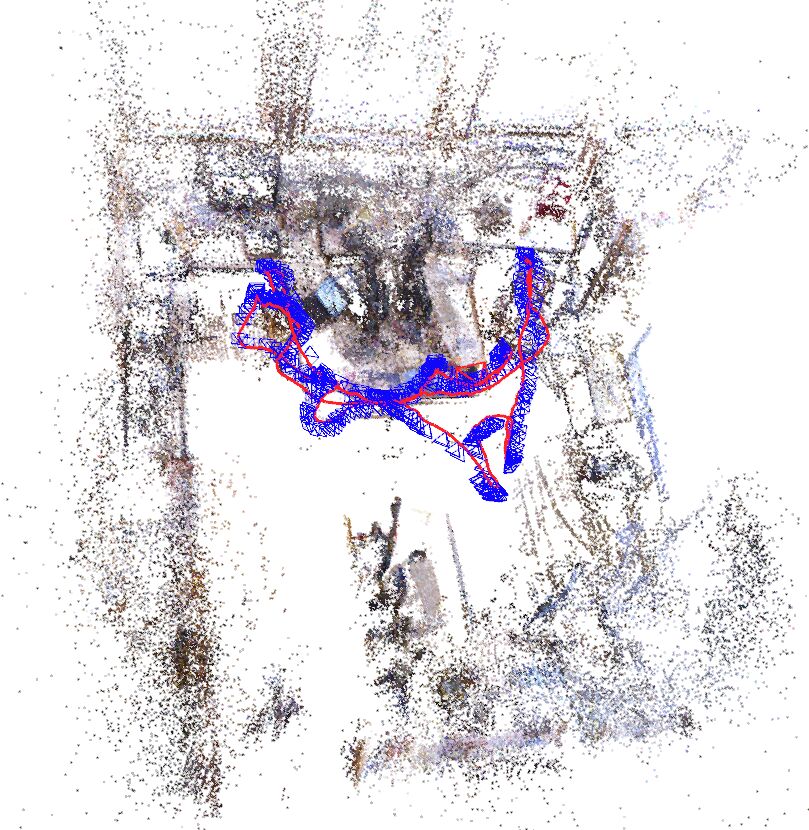}
    \end{overpic}&
    \begin{overpic}[width=.33\columnwidth]{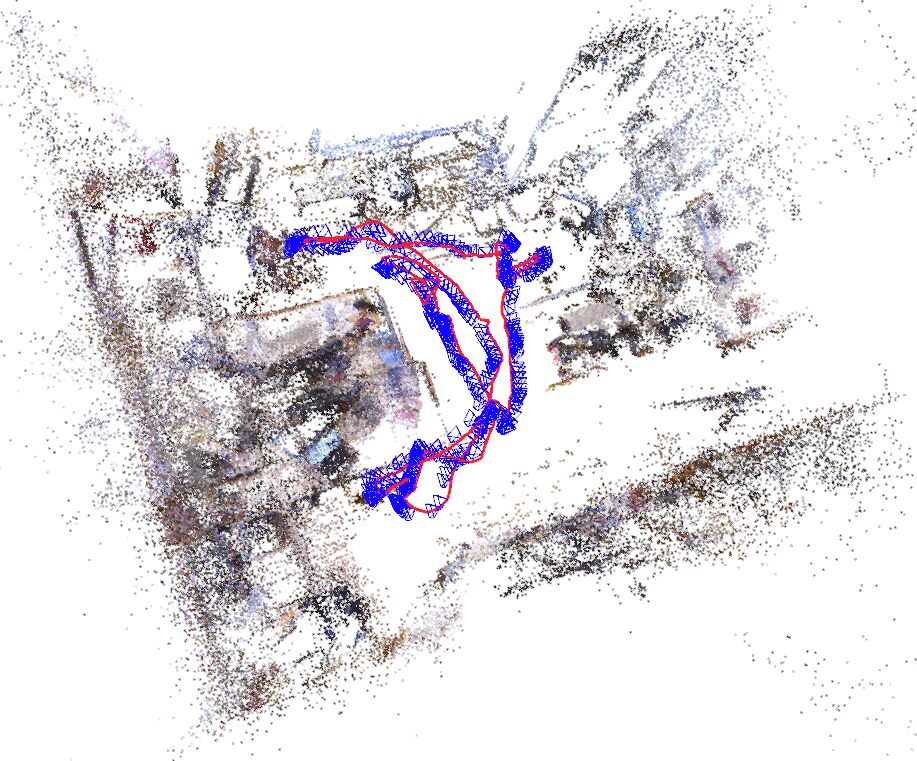}
    \end{overpic}&
    \begin{overpic}[width=.33\columnwidth]{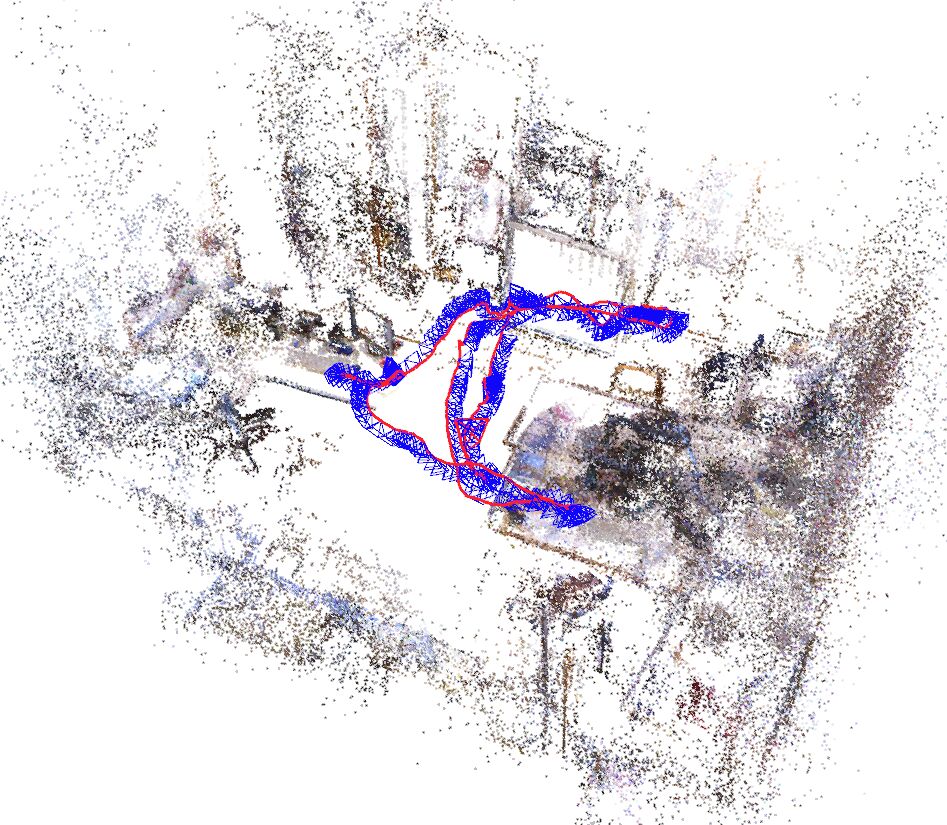}
    \end{overpic}\\
  %%%%%%%%%%%%%%%%%%%%%%%%%%
    \begin{overpic}[width=.33\columnwidth]{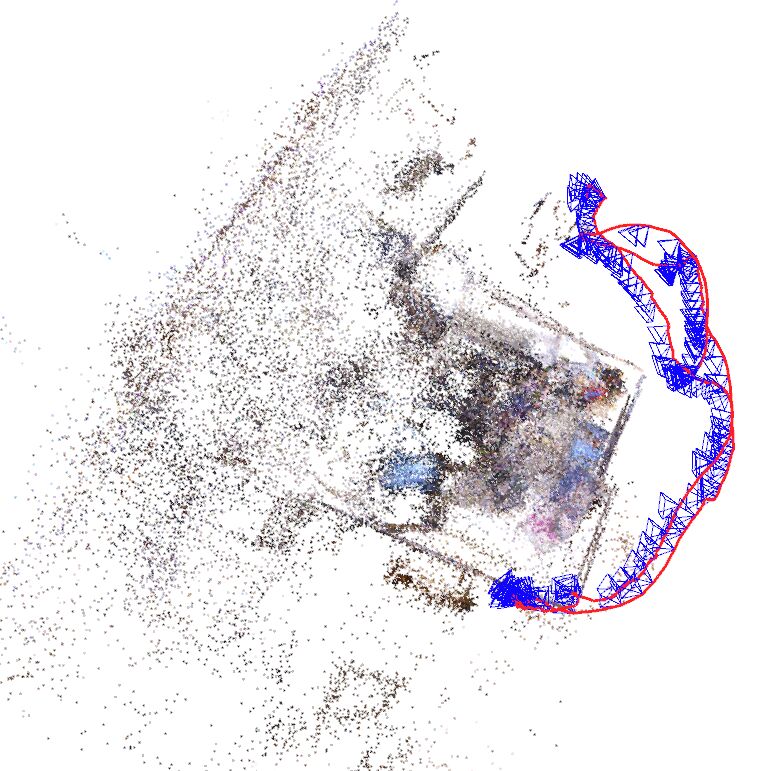}
        \put(6,95){\color{black}\scriptsize\textbf{a) fr1/room}}
        \put(6,-13){\color{black}\scriptsize\textbf{b) fr1/desk2}}
    \end{overpic}&
    \begin{overpic}[width=.33\columnwidth]{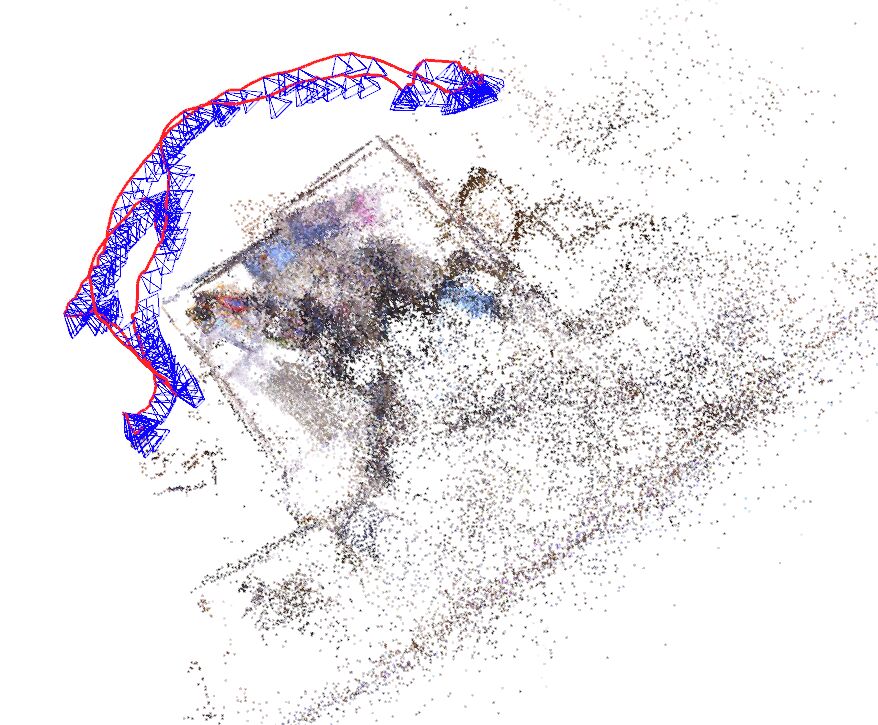}
    \end{overpic}&
    \begin{overpic}[width=.33\columnwidth]{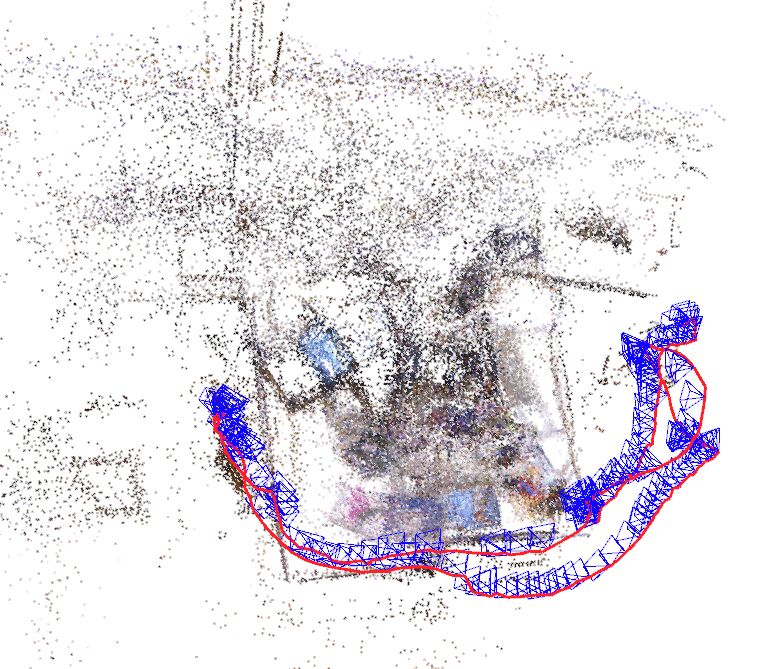}
    \end{overpic}\\
    %%%%%%%%%%%%%%%%%%%%%%%%%%
  \end{tabular}
\end{center}
%\vspace{.cm}
\caption{Examples of localisation and mapping results using our loop closure approach on TUM-RGBD dataset.
The trajectory is shown in red and the camera poses in blue.}
\label{fig:qualitative_recons_tum}
\end{figure}
% ********************************

% ********************************
\begin{figure}[t]
\begin{center}
  \begin{tabular}{@{}c@{}c@{}c}
    \begin{overpic}[width=.33\columnwidth]{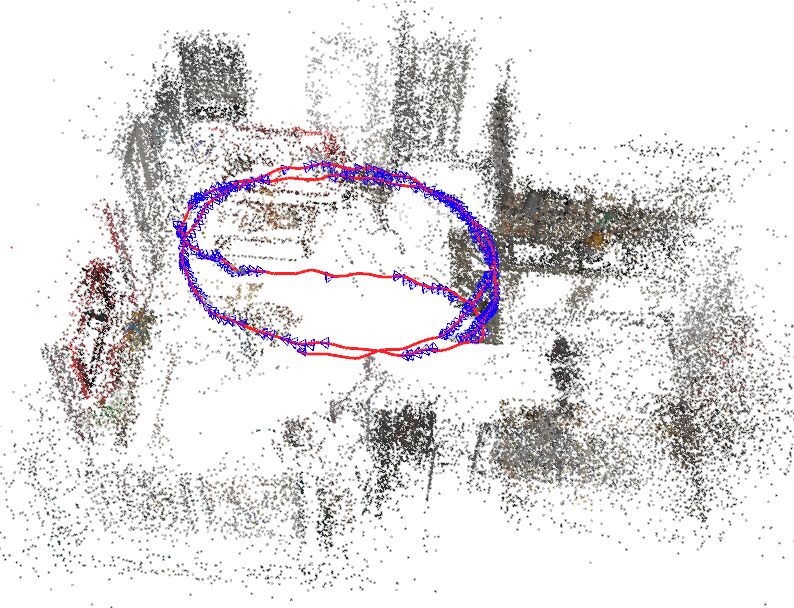}    
    \end{overpic}&
    \begin{overpic}[width=.33\columnwidth]{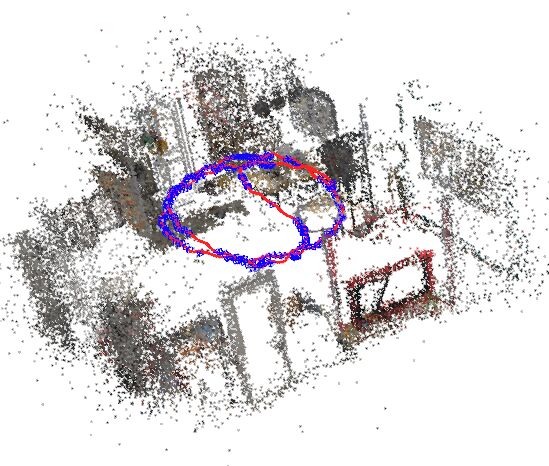}
    \end{overpic}&
    \begin{overpic}[width=.33\columnwidth]{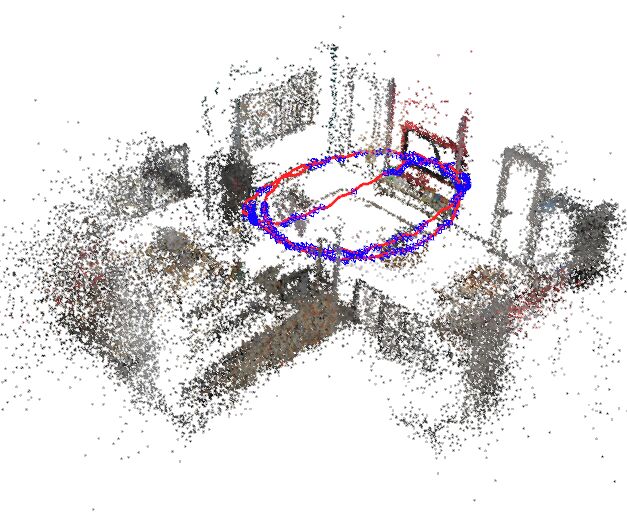}
    \end{overpic}\\
  %%%%%%%%%%%%%%%%%%%%%%%%%%
    \begin{overpic}[width=.33\columnwidth]{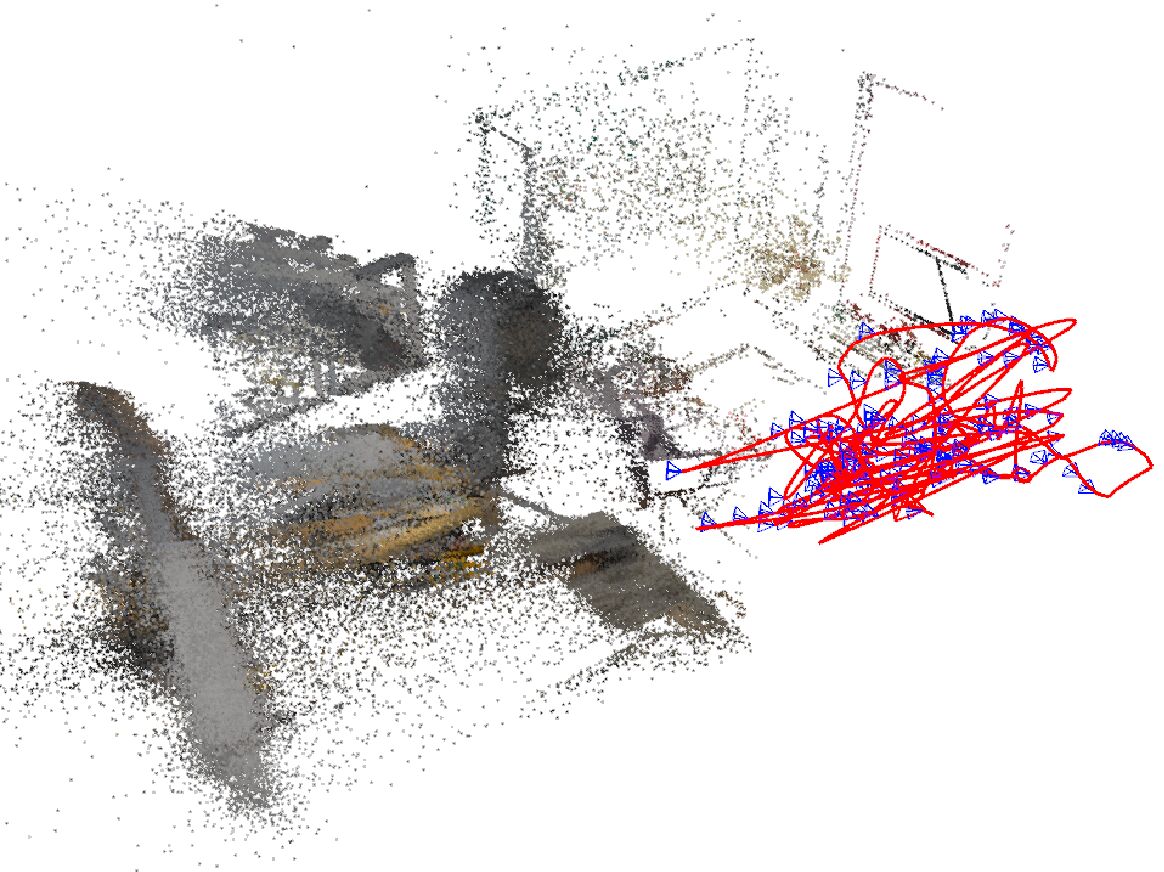}
        \put(6,75){\color{black}\scriptsize\textbf{a) deer/run}}
        \put(6,-7){\color{black}\scriptsize\textbf{b) diamond/Mfast}}
    \end{overpic}&
    \begin{overpic}[width=.33\columnwidth]{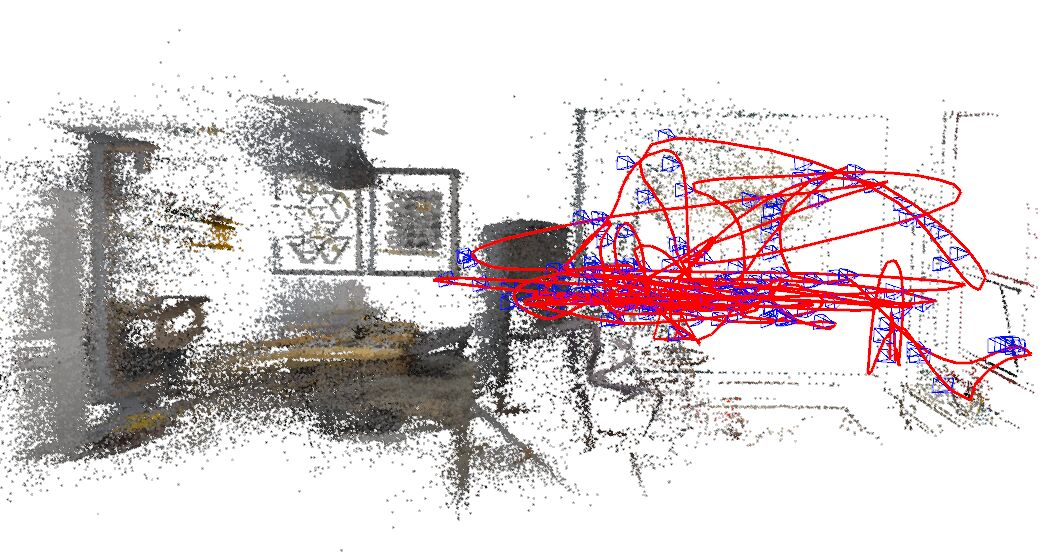}
    \end{overpic}&
    \begin{overpic}[width=.33\columnwidth]{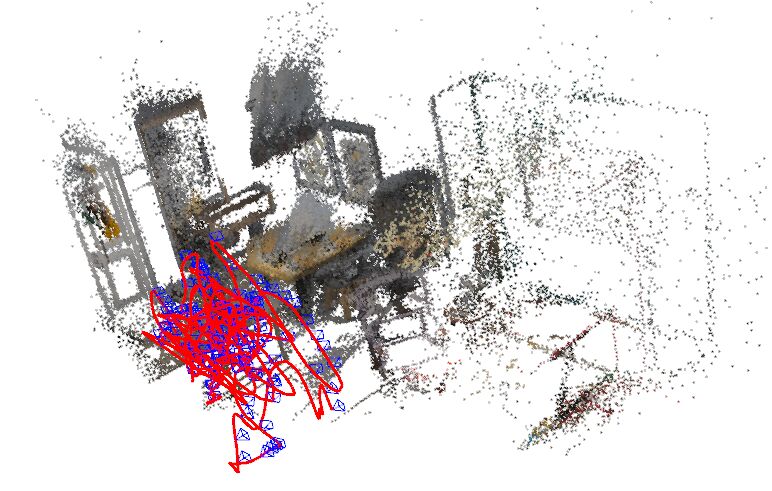}
    \end{overpic}\\
    %%%%%%%%%%%%%%%%%%%%%%%%%%
  \end{tabular}
\end{center}
\vspace{-.1cm}
\caption{Examples of localisation and mapping results using our loop closure approach on ICL dataset.
The trajectory is shown in red and the camera poses in blue.
}
\label{fig:qualitative_recons_icl}
\end{figure}
% ********************************

%%%%%%%%%%%%%%%%%%%%%%%%%%%%%%%%%%%%%%%%%%%%%%%%%%%%%%%%%%%%%%%%%%%%
\noindent\textbf{Discussion.}
Tab.~\ref{tab:quant_results} and Tab.~\ref{tab:quant_results1} report the results of our approach where we can observe that for the majority of the sequences we can improve the localisation accuracy of RESLAM.
We run our algorithm multiple times on a subset of sequences of Tab.~\ref{tab:quant_results} and Tab.~\ref{tab:quant_results1} to assess the numerical stability of the results and we obtained stable results across different runs.
We can observe a reduced RMSE in almost all sequences (except for \texttt{\footnotesize fr1/xyz}) when we use L3D descriptors only to estimate the transformation between the keyframes forming loops (i.e.~RESLAM [w.~L3D]). On average, when comparing RESLAM [w.~L3D] to RESLAM [repro. 1], we obtain $9.5\%$ and $22.4\%$ reduction in RMSE on RGBD-TUM and ICL, respectively.  
%Except for \texttt{\footnotesize fr1/xyz}, where the difference with respect to RESLAM [repro.1] is small, we lower the ATE in all the other sequences.
By employing our loop detection approach, we can observe a further reduction of RMSE. 
On average, when comparing OURS to RESLAM [repro. 1], we obtain $19\%$ and $46.6\%$ reduction in RMSE on RGBD-TUM and ICL, respectively.  
This is mainly due to the detection of a larger number of loops, thus a more frequent triggering of pose graph optimisation which reduces the trajectory error.

We show in Fig.~\ref{fig:qualitative_res_traj} some qualitative results on the RGBD-TUM and ICL datasets. 
These sequences include a variety of trajectory patterns, i.e.~from the handheld camera of \texttt{\footnotesize fr1/room} to the micro aerial vehicle mounted camera of \texttt{\footnotesize deer/run}.
We can visually appreciate how the trajectory estimated using our proposed L3D-based loop closure approach is much closer to the ground-truth one.
Some part of these trajectories are marked and zoomed-in with a purple bounding box to highlight the improvements.

Fig.~\ref{fig:qualitative_res_pc_registration} shows some registration results on the keyframes corresponding to detected loops when the 6DoF transformation is estimated using the RESLAM's loop closure approach and ours.
We can observe that the registration that is obtained with RESLAM provides a systematic misalignment.
We noticed this also on other sequences.
At times, the transformation estimated with RESLAM is incorrect, see c) and d), while our loop closure approach can often register the keyframes correctly.
Lastly, Fig.~\ref{fig:qualitative_recons_tum} and Fig.~\ref{fig:qualitative_recons_icl} show qualitative localisation and mapping results using our loop closure approach.
From these examples we can observe the full structure of the environment in the form of coloured point cloud.

%%%%%%%%%%%%%%%%%%%%%%%%%%%%%%%%%%%%%%%%%%%%%%%%%%%%%%%%%%%%%%%%%%%%
%%%%%%%%%%%%%%%%%%%%%%%%%%%%%%%%%%%%%%%%%%%%%%%%%%%%%%%%%%%%%%%%%%%%
\subsection{Analysis of 3D descriptors}
\label{sec:descriptor_analysis}
We provide a quantitative analysis of traditional and recent 3D deep descriptors using the TUM-RGBD dataset. 
This experiment motivates the choice of DIP descriptor by comparing it with descriptors in the literature.
We adopt the registration recall as the performance measure \cite{Gojcic2019,Poiesi2021}, which measures the ratio of point cloud pairs whose average distance error between the corresponding points after being registered by the estimated transformation is below 0.2m.

We use the edge-based point clouds produced by RESLAM from four sequences of TUM-RGBD. 
For each sequence we randomly select 1000 point cloud pairs with at least 30\% overlap.
We compare four descriptors, one handcrafted approach, namely FPFH \cite{rusu2009fast}, and three deep learning based approaches: one global approach, namely FCGF \cite{Choy2019a}, and two local approaches, namely SpinNet \cite{Ao2021} and DIP \cite{Poiesi2021}.
For FPFH we used its Open3D implementation, whereas for the others, we used the code and models provided by the authors.
For all the deep learning based descriptors, we use their models trained on the 3DMatch dataset \cite{Zeng2017}.

Tab.~\ref{tab:desc_choice} shows the results of this experiment.
FPFH and FCGF are the fastest descriptors to compute, however at a large cost of the registration error.
DIP achieves the best registration recall with a better compromise between the computation efficiency and the registration performance compared to SpinNet.
%We decided to use the descriptor with the best recall performance in our experiments, as computation time can be improved via code optimisations.

%+++++++++++++++++++++++++
\begin{table}[t]
    \tabcolsep 2pt
    \centering
    \caption{Registration recall and computation time per descriptor in [ms] computed on four sequences of RGBD-TUM dataset.}
    \label{tab:desc_choice}
    % \resizebox{1\columnwidth}{!}{%
    \vspace{-.2cm}
    \begin{tabular}{lcccc|cc}
        \toprule
         & \texttt{\scriptsize fr1/desk} & \texttt{\scriptsize fr1/desk2} & \texttt{\scriptsize fr1/plant} & \texttt{\scriptsize fr1/room} & mean & [ms] \\
        \midrule
        FPFH \cite{rusu2009fast} & 0.35 & 0.39 & 0.34 & 0.41 & 0.37 & 0.02 \\
        FCGF \cite{Choy2019a} & 0.58 & 0.46 & 0.41 & 0.52 & 0.50 & 0.03 \\
        SpinNet \cite{Ao2021} & 0.71 & 0.75 & \textbf{0.85} & 0.75 & 0.76 & 10.18 \\
        DIP \cite{Poiesi2021} & \textbf{0.73} & \textbf{0.76} & \textbf{0.85} & \textbf{0.77} & \textbf{0.78} & 3.79 \\
        \bottomrule
    \end{tabular}
\end{table}
%+++++++++++++++++++++++++

\section{Conclusions}
We presented a novel approach to address loop closures using deep learning-based 3D local descriptors \cite{Poiesi2021}.
Our approach detects loop candidates using the ratio of mutually nearest-neighbour descriptors and confirms their quality by computing the novel measure RON, the ratio of metrically near points amongst points that are mutually nearest neighbours in the descriptor space.
We evaluated the approach for both LiDAR-based and visual SLAM system.
Results showed that our approach can outperform the state-of-the-art LiDAR-based loop closure detection methods, and further improves the localisation accuracy of the visual SLAM system. 
Future research directions include fusing 2D visual cue extracted from images with 3D local descriptors to handle scenes lacking geometric structures, and focus on the optimisation of descriptor computation.
%We will also analyse the instability of pose graph optimisation more thoroughly.

%%%%%%%%%%%%%%%%%%%%%%%%%%%%%%%%%%%%%%%%%%%%%%%%%%%%%%%%%%%%%%%%%%%%%%%%%%%%%%%%
\bibliographystyle{IEEEtran}
\bibliography{refs}

\end{document}